\RequirePackage{fix-cm}

\documentclass[twocolumn]{svjour3}
\usepackage{empheq}

\smartqed  

\usepackage{graphicx}%

\usepackage{amsmath}
\usepackage{algorithm}
\usepackage{algpseudocode}  
\usepackage{amsfonts}
\usepackage{enumerate}
\usepackage[accsupp]{axessibility}

\usepackage{empheq}
\usepackage{natbib}
\usepackage{rotating}
\usepackage{ragged2e}
\usepackage{appendix}
\usepackage{booktabs}
\usepackage{color}
\usepackage{soul}
\usepackage{colortbl}  
\usepackage{multirow}
\usepackage{bm}
\usepackage{bbm}
\usepackage{caption}
\usepackage[labelformat=simple]{subcaption}
\usepackage{hyperref}
\hypersetup{
    colorlinks=true,
    linkcolor=red,
    filecolor=blue,      
    urlcolor=blue,
    citecolor=blue,
}
\usepackage[misc]{ifsym}
\usepackage{multirow}
\usepackage[table,xcdraw]{xcolor}
\usepackage{overpic}
\usepackage{bbding}
\newcommand{\etal}{\textit{et~al}.~}
\newcommand{\ie}{\textit{i}.\textit{e}.,~}
\newcommand{\eg}{\textit{e}.\textit{g}.,~}
\definecolor{skyblue}{HTML}{03FFFF}
\definecolor{lightyellow}{HTML}{FFFF00}
\definecolor{lightgreen}{HTML}{03FF00}
\definecolor{lightpink}{HTML}{FFE2E1}
\definecolor{lightblue}{HTML}{c9e5ff}
\definecolor{lightpurple}{HTML}{abacf4}
\definecolor{lightorange}{HTML}{ffcc67}
\definecolor{lightgray}{HTML}{c0c0c0}
\definecolor{lightred}{HTML}{FF8181}

\usepackage{colortbl}
\usepackage{booktabs}

\definecolor{Gray1}{rgb}{0.92,0.92,0.92}
\definecolor{Gray2}{rgb}{0.88,0.88,0.88}

\begin{document}\sloppy

\title{IPDiff: Diffusion-driven ORSI Salient Object Detection with Information Reconstruction and Multi-Prior Guidance}


\author{Gongyang~Li\textsuperscript{1}\and
        Zhen~Bai\textsuperscript{2}\and
        Runmin~Cong\textsuperscript{3}\and
        Dan~Zeng\textsuperscript{1}\and
        Weisi~Lin\textsuperscript{4}\and
        Xiao-Ping~Zhang\textsuperscript{5}
}

\authorrunning{Li,~\emph{et al.}} 

\institute{
\Letter~~Runmin Cong\\
\email{rmcong@sdu.edu.cn}\\\\
\Letter~~Dan Zeng\\
\email{dzeng@shu.edu.cn}\\\\
Gongyang Li\\
\email{ligongyang@shu.edu.cn}\\\\
Zhen Bai\\
\email{bz536476@163.com}\\\\
Weisi Lin\\
\email{wslin@ntu.edu.sg}\\\\
Xiao-Ping Zhang\\
\email{xpzhang@ieee.org}\\\\
\textsuperscript{1} School of Communication and Information Engineering, Shanghai University, Shanghai, China \\\\
\textsuperscript{2} Department of Medical Equipment, the First Affiliated Hospital of Zhengzhou University, Zhengzhou, China\\\\
\textsuperscript{3} School of Control Science and Engineering, Shandong University, Jinan, China\\\\
\textsuperscript{4} School of Computer Science and Engineering, Nanyang Technological University, Singapore\\\\
\textsuperscript{5} Shenzhen Key Laboratory of Ubiquitous Data Enabling, Tsinghua Shenzhen International Graduate School, Tsinghua University, Shenzhen, China\\\\
}

\date{Received: date / Accepted: date}

\maketitle
\begin{abstract}
%
%
Existing Salient Object Detection in Optical Remote Sensing Image (ORSI-SOD) methods mainly adopt the static inference strategy, which uses fixed trained model parameters for saliency inference in the testing phase.
This means that even if the generated saliency map has errors, it cannot be further optimized.
%
In this paper, we propose the novel \emph{IPDiff}, a \underline{\emph{Diff}}usion-driven ORSI-SOD method with \underline{\emph{I}}nformation Reconstruction and Multi-\underline{\emph{P}}rior Guidance.
We build IPDiff based on a unique dynamic optimization strategy, which endows IPDiff with the ability to iteratively optimize saliency maps with a dynamic parameter.
Specifically, we formulate ORSI-SOD as a conditional diffusion problem in IPDiff. 
IPDiff first extracts informative conditional priors from ORSIs, including the saliency prior and the hierarchical priors, in the prior network with the assistance of the information reconstruction-driven attention module.
The saliency prior can provide positional information of salient objects, while the hierarchical priors can provide specific detail and semantic information of salient objects.
Under the guidance of these priors, IPDiff then iteratively denoises random noise as the timestep dynamically changes in the denoising network, generating saliency maps that are close to ground truths.
%
Notably, we simultaneously supervise IPDiff in both spatial and spectral domains through a hybrid loss function to achieve efficient network training. 
Comprehensive experiments on public ORSSD, EORSSD, and ORSI-4199 datasets demonstrate that our proposed IPDiff achieves the best performance compared to 46 state-of-the-art methods.
The code and results of our method are available at https://github.com/MathLee/IPDiff.

\end{abstract}
\section{Introduction}
\label{sec:introduction}

%
Optical Remote Sensing Images (ORSIs) refer to images captured by cameras mounted on spacecraft, drones, airplanes, \textit{etc}, and have the characteristics of strong intuitiveness, high spatial resolution, and rich spectral information~\citep{2025IJCVLSKNet}.
ORSI processing plays an important role in various fields, such as agriculture, military, oceanography, ecological protection, and geological exploration~\citep{2025NMI}.
Salient Object Detection (SOD) aims to pop out the most attractive objects in images or videos~\citep{2017IJCVSOD,2021IJCVSODTPT,2021IJCVRGBDSOD,2022IJCVRGBDMFAS,2023IJCVRGBDDCD,2024IJCVRGBDCMFPD,2024IJCVViDSOD,2026WSOD,2026TSD,DiffusionRGBD,26STENet,2026FreMaNet,2023SGFusion}.
It is a fundamental topic in the computer vision community.
Recently, ORSI-SOD~\citep{2019LVNet,2021DAFNet,2022MJRBM,2023GeleNet,2024PRNet,2025MRBINet,2025DPUFormer} has become a hot topic.
It can quickly locate eye-catching objects in ORSIs, and serves as the cornerstone for ORSI understanding and interpretation.

\begin{figure}[t!]
  \centering
  \footnotesize
  \begin{overpic}[width=1\columnwidth]{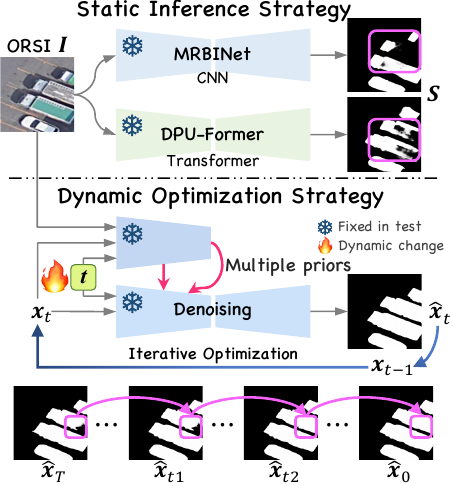}
  \end{overpic}
  \caption{%
  Two strategies in ORSI-SOD.
  Existing static inference strategy uses fixed trained model parameters
for saliency inference, 
  which is formulated as $\boldsymbol{S}={\phi}_{\theta}(\boldsymbol{I})$ with fixed trained parameters $\theta$.
  ${\phi}(\cdot)$ represents the ORSI-SOD model, such as CNN-based MRBINet~\citep{2025MRBINet} and transformer-based DPU-Former~\citep{2025DPUFormer}.
  Our unique dynamic optimization strategy iteratively optimizes saliency maps as $t$ dynamically changes, 
  which is formulated as $\widehat{\boldsymbol{x}}_{t}={\psi}_{\theta}(\boldsymbol{x}_{t},t,\boldsymbol{I}) \& \widehat{\boldsymbol{x}}_t \stackrel{\text{Sample}}{\longrightarrow} \boldsymbol{x}_{t-1}$ and $\boldsymbol{S}=\widehat{\boldsymbol{x}}_{0}$ with fixed trained parameters $\theta$ and dynamically changing $t$.
  ${\psi}(\cdot)$ represents our IPDiff.
    }
  \label{fig:strategies}
\end{figure}

ORSI-SOD has made breakthroughs in succession with the assistance of deep learning technologies, such as Convolutional Neural Networks (CNNs)~\citep{2015VGG,2016ResNet} and transformers~\citep{2022PVTv2,2021Swin}.
The existing ORSI-SOD methods can be divided into four categories, 
including CNN-based methods~\citep{2019LVNet,2021SARNet,2021DAFNet,2022EMFINet,2022MJRBM,2022MCCNet,2022ERPNet,2023ACCoNet,2024PRNet,2024MIRGNet,2024TSCNet,2024RAGRNet,2024SFANet,2025BCARNet,2025MRBINet},
transformer-based methods~\citep{2023GLGCNet,2023GeleNet,2024TLCKDNet,2024UDCNet,2025DASGNet,2025DPUFormer,2025PEFFNet,2025DKETFormer,2025LBAMCNet}, 
hybrid backbone-based methods~\citep{2022HFANet,2024ADSTNet,2025PIFRNet,2025TSFANet}, and 
lightweight methods~\citep{2022CorrNet,2023SeaNet,2024SAFINet,2024LPMFCNet,2025SggNet,2025RAMENet,2025SOLNet}.
As the name implies, the CNN-based methods typically adopt CNNs, such as VGG~\citep{2015VGG} and ResNet~\citep{2016ResNet}, as backbones, and explore edge cues~\citep{2022EMFINet,2022MJRBM,2024RAGRNet}, multi-level interaction~\citep{2021DAFNet,2023ACCoNet,2024MIRGNet}, and multi-input architecture~\citep{2019LVNet,2022EMFINet}.
The transformer-based methods typically adopt transformers, such as pyramid vision transformer~\citep{2022PVTv2} and Swin Transformer~\citep{2021Swin}, as backbones, and explore the global-local-global scheme~\citep{2023GLGCNet}, the global-to-local paradigm~\citep{2023GeleNet,2025DASGNet}, and the global-local integration strategy~\citep{2024UDCNet,2025DPUFormer,2025LBAMCNet}.
While hybrid backbone-based methods simultaneously use CNNs and transformers.
Different from the above three categories, the lightweight methods aim to achieve a balance between performance and model complexity for practical applications.

%
%
%
%
The above four categories of ORSI-SOD methods can be summarized as the predictive framework, mainly training a prediction model to learn to infer saliency values from a large amount of ORSI data.
After training, they adopt the static inference strategy for saliency inference, \ie using fixed trained model parameters to infer saliency maps once.
This strategy has an obvious drawback, \ie the erroneous saliency values once occurring cannot be corrected.
As shown in Fig.~\ref{fig:strategies}, the erroneous region (pink boxes) in the saliency maps generated by CNN-based MRBINet~\citep{2025MRBINet} and transformer-based DPU-Former~\citep{2025DPUFormer} will persist.
Different from the predictive framework in ORSI-SOD, researchers develop the generative adversarial framework for SOD in Natural Scene Images (NSI-SOD)~\citep{2018SODGAN,2020SODGAN}.
In the training phase, they train both the generative model and the discriminative model to improve the ability of the generative model to generate high-quality saliency maps.
However, after training, they use the generative model with fixed training parameters to infer saliency values once, which also belongs to the static inference strategy.
In addition, the generative adversarial framework has training instability and mode collapse issues, which limit its application in SOD.



%
%

The generative adversarial framework is eye-catching, inspiring us to explore a generation solution for ORSI-SOD.
We have noticed that recent diffusion models~\citep{2020DDPM,2021IDDPM,2021DDIM,2022StableDiffusion,2023SimpleDiffusion} may be a suitable generative solution.
Diffusion models are originally intended for image and video generation topics.
Its core principle lies in simulating the processes of noise diffusion and reverse denoising to generate realistic data from random noise.
Moreover, it controls the generation process through dynamic timesteps.
Therefore, we attempt to solve ORSI-SOD from a generative perspective based on the diffusion model, and propose a dynamic optimization strategy to break through the limitations of the previous static inference strategy.
The strategy performs denoising with fixed trained parameters and dynamically changing timesteps in the testing phase, which means \emph{the saliency inference is no longer a one-time occurrence}.
With this strategy, we propose the novel IPDiff, a diffusion-driven ORSI-SOD method with information reconstruction and multi-prior guidance.
As shown in Fig.~\ref{fig:strategies}, the dynamic optimization strategy endows our IPDiff with the ability to iteratively optimize saliency maps, \ie the missing truck carriage in the saliency map is completely segmented out with the dynamic change of timesteps.


Unlike the vanilla diffusion model that directly generates images from noise, we regard ORSI-SOD as a conditional diffusion problem.
Our IPDiff treats ORSIs as condition information to guide denoising from noise.
In particular, IPDiff consists of a prior network and a denoising network.
The prior network extracts specific conditional priors from ORSIs, including the saliency prior and the hierarchical priors.
To extract the useful content of priors, we propose the Information Reconstruction-driven Attention Module (IRAM) to adaptively reconstruct features in the spectral domain.
Subsequently, these multiple priors are sequentially injected into the denoising network.
In the denoising network, the saliency prior stabilizes the position of salient objects, and then the hierarchical priors hierarchically enrich the details and semantics of salient objects.
To mitigate the negative impact of the noisy mask, we propose the Information Perturbation Module (IPM) to enhance the anti-interference capability of the denoising network.
Notably, IPM is only equipped in the training phase.
With all components working together, our IPDiff can generate accurate saliency maps through iterative optimization as the timestep dynamically changes, showing good adaptation to the complex and variable scenes of ORSIs.

Our main contributions are summarized as follows:
\begin{itemize} 
\item We propose a novel diffusion-driven ORSI-SOD framework based on the unique dynamic optimization strategy, namely \emph{IPDiff}, which differs from previous methods based on the static inference strategy.
IPDiff formulates ORSI-SOD as a conditional diffusion problem, which first extracts conditional priors from ORSIs as guidance, and then iteratively denoises random noise to generate saliency maps close to ground truths.

\item We propose the IRAM to achieve robust enhancement on basic features, generating informative conditional priors.
IRAM reconstructs information in the spectral domain through adaptive spectrum decoupling and information aggregation, and produces the attention map from the reconstructed information to enhance features.

\item We propose a Multi-Prior Guidance Denoising Network to optimize saliency maps step-by-step by denoising random noise under the guidance of the saliency prior and the hierarchical priors.
Notably, our denoising network has strong feature representation and anti-interference capabilities, as it is equipped with multiple IPMs in the training phase.
\end{itemize}

\section{Related Work}
\label{sec:related}

\subsection{Salient Object Detection in Optical Remote Sensing Images}
\label{sec:ORSI_SOD}

Recently, ORSI-SOD has developed rapidly and occupies an important position in the field of SOD.
This is attributed to the repeated breakthroughs in deep learning technologies, such as CNNs~\citep{2015VGG,2016ResNet}, transformers~\citep{2022PVTv2,2021Swin}, and attention mechanisms~\citep{2018CBAM,2017transformer}.
With the application of various technologies, the challenges of ORSI-SOD have been overcome one by one, and its performance has gradually improved.
At present, existing ORSI-SOD methods can be divided into four categories.
The first three categories are classified according to the differences in the backbones used, namely CNN-based methods, transformer-based methods, and hybrid backbone-based methods.
The last category focuses on the complexity of the model, called lightweight methods.

CNN-based ORSI-SOD methods~\citep{2019LVNet,2021SARNet,2021DAFNet,2022EMFINet,2022MJRBM,2022MCCNet,2022ERPNet,2023ACCoNet,2024PRNet,2024MIRGNet,2024TSCNet,2024RAGRNet,2024SFANet,2025BCARNet,2025MRBINet} account for a large proportion, due to the classic architecture of CNNs (\ie VGG~\citep{2015VGG} and ResNet~\citep{2016ResNet}) and their powerful feature extraction capabilities.
In this category, researchers employed edge cues to outline the boundaries of salient objects, improving the fine-grained details of salient objects~\citep{2022ERPNet,2022EMFINet,2022MCCNet,2022MJRBM,2024RAGRNet,2025MRBINet}.
While Huang~\etal\citep{2021SARNet} and Quan~\etal\citep{2024SFANet} exploited high-level semantic cues to accurately locate salient objects and reduce omissions of salient objects.
The multi-level interaction is a popular strategy.
It aims to explore the complementarity of information at different granularities among multi-level features of various scales to better represent salient objects~\citep{2021DAFNet,2023ACCoNet,2024MIRGNet,2024TSCNet,2025BCARNet,2024PRNet}.
Different from the above methods, Li~\etal\citep{2019LVNet} and Zhou~\etal\citep{2022EMFINet} used the multi-input architecture to directly extract multi-scale features from multiple ORSIs of different scales to adapt to complex ORSI scenes.

Transformer-based ORSI-SOD method~\citep{2023GLGCNet,2023GeleNet,2024TLCKDNet,2024UDCNet,2025DASGNet,2025DPUFormer,2025PEFFNet,2025DKETFormer,2025LBAMCNet} is a rising star.
It utilizes transformers~\citep{2022PVTv2,2021Swin} to establish long-range dependencies of ORSIs, extracting the global information of ORSIs.
Based on the global information, the global-local-global scheme~\citep{2023GLGCNet}, the global-to-local paradigm~\citep{2023GeleNet,2025DASGNet}, and the global-local integration strategy~\citep{2024UDCNet,2025DPUFormer,2025LBAMCNet} have been proposed successively to interact global and local information from different perspectives.
These methods overcame the limitations of CNN-based methods which cannot model global information, further improving detection performance.
Moreover, to address the problem of complex orientations in ORSIs, researchers extracted directional cues from global information by using convolutions with multiple directions to better perceive and determine the directions of salient objects~\citep{2023GeleNet,2025DASGNet,2025DKETFormer}.


Hybrid backbone-based methods~\citep{2022HFANet,2024ADSTNet,2025PIFRNet,2025TSFANet} draw on the advantages of different types of backbones and jointly extract comprehensive features.
There are three ways to integrate different types of backbones.
The first way integrates CNN blocks and transformer blocks in a single-stream structure.
Wang~\etal\citep{2022HFANet} followed the first way to model local and global context at different levels from ORSIs.
The second way uses CNN and transformer in parallel in a dual-stream structure.
Zhao~\etal\citep{2024ADSTNet} and Wang~\etal\citep{2025PIFRNet} followed the second way to extract local and global information and fuse them.
The last way is also a dual-stream structure.
The first stream is the same as the first way, and the second stream integrates CNN blocks and Mamba blocks~\citep{VMamba}.
Li~\etal\citep{2025TSFANet} developed the complex backbone to leverage the global modeling capabilities of transformer and the local processing advantages of Mamba, enriching the feature extraction manner of ORSI-SOD.







Lightweight ORSI-SOD methods~\citep{2022CorrNet,2023SeaNet,2024SAFINet,2024LPMFCNet,2025SggNet,2025RAMENet,2025SOLNet} generally use lightweight backbones and develop lightweight modules to achieve efficient SOD.
As a pioneer, Li~\etal\citep{2022CorrNet} modified the vanilla VGG to a lightweight VGG backbone, greatly reducing model parameters to 4.09M.
Subsequently, MobileNet series~\citep{MobileNet2,MobileNet3} dominated the lightweight  ORSI-SOD category.
MobileNet-V2~\citep{MobileNet2} was combined with the multi-level collaboration strategy~\citep{2023SeaNet,2025SggNet} and the attention recursion mechanism~\citep{2024SAFINet}, further reducing the model parameters and the computational load.
Cheng~\etal\citep{2024LPMFCNet} combined MobileNet-V3~\citep{MobileNet3} with the multi-level collaboration strategy, achieving competitive performance with 0.5G FLOPs.
Differently, Han~\etal\citep{2025RAMENet} introduced MobileViT~\citep{MobileViT} to enhance the feature extraction, improving performance by increasing a few parameters.
Li~\etal\citep{2025SOLNet} introduced RepVGG structure~\citep{RepVGG} to accelerate the inference speed to 161 fps.

\begin{figure*}
	\centering
	\begin{overpic}[width=1\textwidth]{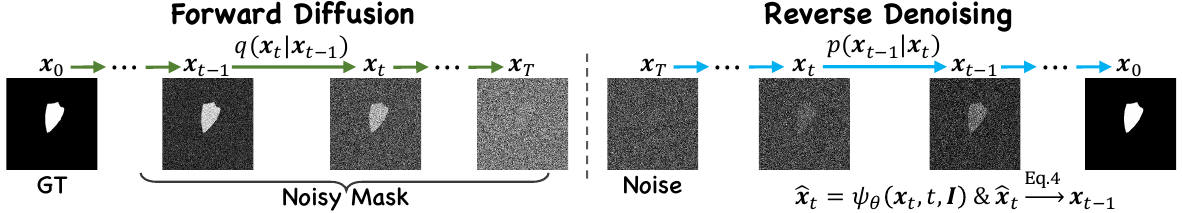}
    \end{overpic}
	\caption{Illustration of the forward diffusion process and the reverse denoising process.
    }
    \label{fig:Preliminaries}
\end{figure*}

%
%
%

Although the four aforementioned categories of methods have greatly promoted the development of ORSI-SOD, they all adopted the static inference strategy for one-time saliency inference.
The inherent drawback of this strategy results in saliency maps inferred by existing methods being unoptimizable.
This places high demands on the design of ORSI-SOD methods, requiring them to have strong adaptability to various scenes of ORSIs so that they can infer satisfactory saliency maps at one time.
Obviously, this is hard.
To break through the current situation, we introduce the dynamic optimization strategy into our ORSI-SOD solution, \ie IPDiff.
The dynamic optimization strategy can perform multiple iterative inferences in the testing phase based on dynamically changing timesteps and fixed trained parameters.
The erroneous saliency values in the previous inference can be corrected in the current inference.
In this way, our IPDiff can continuously optimize saliency maps through multiple iterative inferences, being able to generate accurate saliency maps.

\subsection{Denoising Diffusion Models}
\label{sec:Diffusion}

Denoising diffusion model is inspired by nonequilibrium thermodynamics~\citep{2020DDPM}.
It learn the data distribution by simulating the two-way process of \emph{gradual addition of noise} and \emph{step-by-step removal of noise}, and ultimately generates realistic data from random noise, such as images and videos.
Since its proposal, diffusion models have been optimized in multiple aspects, such as sampling acceleration, generation speed, synthesis resolution, scalability, and text-to-image generation~\citep{2021IDDPM,2021DDIM,2022StableDiffusion,2023SimpleDiffusion}.




The powerful generation capability of diffusion models has led to their application in some visual tasks, such as aerial semantic segmentation~\citep{2024SatSynth}, cardiac ultrasound segmentation~\citep{2025UltrasoundSeg}, steel surface defect detection~\citep{2025DefFiller}, and fixation prediction~\citep{2024SalPre}.
Researchers promoted performance improvement by synthesizing training data through diffusion models, \ie using diffusion models for data augmentation.
Obviously, these methods face the problems of data quality and data authenticity.
We believe that the application of diffusion models in the above visual topics has not deviated from the data generation capability inherent to diffusion models.


Differently, some researchers used diffusion models to treat visual tasks, such as medical image segmentation~\citep{2022EnsemDiff,2023MedSegDiff,2024MedSegDiffv2}, edge detection~\citep{2024DiffusionEdge}, change detection~\citep{2024GCDDDPM}, and camouflaged object detection~\citep{2025CamoDiffusion}, as image-to-image translation tasks.
They modified the vanilla diffusion framework to a conditional diffusion framework, that is, extracting conditional information from inputs and inserting it into the denoising process to generate corresponding outputs.
This way of using diffusion models provides fresh research directions for the computer vision community.
The research focus will be on \emph{how to effectively utilize conditional information to guide the denoising process} or \emph{how to design an effective denoising network}.


Since the diffusion model has the ability to optimize the output according to the timestep, we handle ORSI-SOD with the diffusion model in this paper.
Considering existing issues in ORSI-SOD, we regard ORSI-SOD as a conditional diffusion problem in our IPDiff.
Existing methods based on conditional diffusion for other tasks~\citep{2022EnsemDiff,2023MedSegDiff,2024MedSegDiffv2,2024DiffusionEdge,2024GCDDDPM,2025CamoDiffusion} usually only extract one type of conditional information from the input, which is insufficient.
Differently, we extract multiple conditional priors from the input ORSI in our prior network, including the saliency prior and the hierarchical priors.
The saliency prior is an initial saliency map generated from the input ORSI, and can provide the position information of salient objects to the encoder of our denoising network.
The hierarchical priors are the different scale features of the input ORSI, and can provide specific detail and semantic information of salient objects to the encoder of our denoising network.
Under the guidance of these priors, in our denoising network, the encoder can thoroughly perceive the current noisy mask and effectively guide the decoder to recover salient objects.

\section{Methodology}
\label{sec:OurMethod}
%

\begin{figure*}
	\centering
	\begin{overpic}[width=1\textwidth]{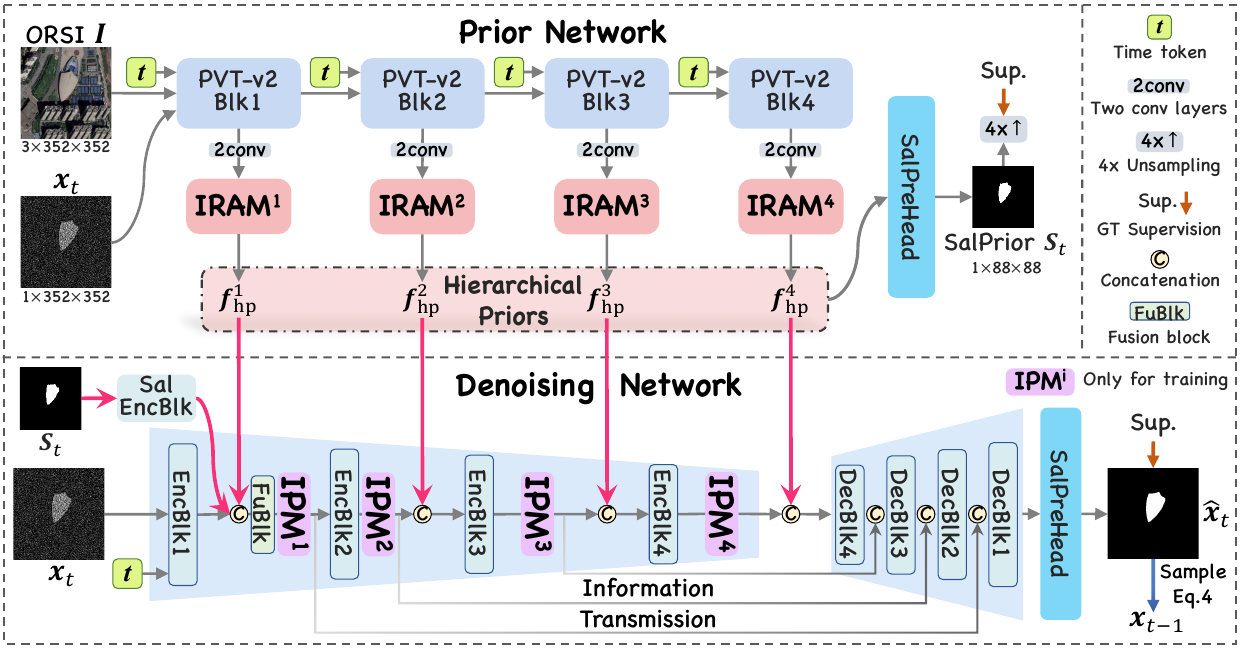}
    \end{overpic}
	\caption{The overall framework of the proposed IPDiff.
    IPDiff consists of a prior network and a denoising network.
    First, the prior network encodes specific conditional priors from ORSIs, including the saliency prior $\boldsymbol{S}_t$ and the hierarchical priors $\{\boldsymbol{f}^{i}_{\rm hp}\}^4_{i=1}$, with the assistance of PVT-v2 and IRAMs.
    Then, under the guidance of these multiple priors, the denoising network denoises the noisy mask $\boldsymbol{x}_t$ to recover the clear one.
    Notably, we only equip IPMs in the encoder of the denoising network in the training phase to enhance the anti-interference capability of the denoising network.
    }
    \label{fig:Framework}
\end{figure*}

\subsection{Preliminaries}
\label{sec:DiffusionPreliminaries}

In general, the denoising diffusion framework~\citep{2020DDPM} includes one process for \emph{gradual addition of noise} (\ie the forward diffusion process) and one process for \emph{step-by-step removal of noise} (\ie the reverse denoising process).
As shown in Fig.~\ref{fig:Preliminaries}, in the forward diffusion process, the ground truth (GT) $\boldsymbol{x}_0$ is gradually noised over the timestep $t\in\{1,2,...,T\}$, obtaining a set of noisy masks $\{\boldsymbol{x}_t\}^{T}_{t=1}$.
On the contrary, in the reverse denoising process, the random Gaussian noise $\boldsymbol{x}_T$ is gradually clear until it is recovered to the original data $\boldsymbol{x}_0$.

\textit{1) Forward Diffusion Process:}
The forward diffusion process is a Markov noising process, producing the noisy mask $\boldsymbol{x}_t$ as follows:
\begin{equation}
   \begin{aligned}
      q(\boldsymbol{x}_t|\boldsymbol{x}_{t-1}) = \mathcal{N}(\boldsymbol{x}_t; \sqrt{{1-\beta}_t} \boldsymbol{x}_{t-1}, {{\beta}_t} \mathbf{I}),
    \label{eq:1}
    \end{aligned}
\end{equation}
where $\mathcal{N}(\cdot)$ is the Gaussian distribution,  $\beta_{t}\in(0,1)$ is the noise variance schedule~\citep{2020DDPM}, $\mathbf{I}$ is the identity matrix, and ${{\beta}_t} \mathbf{I}$ forms the covariance matrix.
%
Therefore, through the propagation of Markov chain, starting from $\boldsymbol{x}_0$ (\ie GT), we can obtain $\boldsymbol{x}_t$ as follows:
\begin{equation}
   \begin{aligned}
      q(\boldsymbol{x}_t|\boldsymbol{x}_0) = \prod_{s=1}^t q(\boldsymbol{x}_s|\boldsymbol{x}_{s-1})= \mathcal{N}\left(\boldsymbol{x}_t; \sqrt{\bar{\alpha}_t} \boldsymbol{x}_0, (1 - \bar{\alpha}_t) \mathbf{I}\right),
    \label{eq:2}
    \end{aligned}
\end{equation}
where $\bar{\alpha}_t = \prod_{s=1}^t \alpha_s = \prod_{s=1}^t (1-\beta_s)$.
Through the above forward diffusion process, we can get a set of noisy masks $\{\boldsymbol{x}_t\}^{T}_{t=1}$ for training.

\textit{2) Reverse Denoising Process:}
The goal of the reverse denoising process is to model the posterior distribution $p(\boldsymbol{x}_{t-1}|\boldsymbol{x}_t)$ as follows:
\begin{equation}
   \begin{aligned}
      p(\boldsymbol{x}_{t-1}|\boldsymbol{x}_t) = \mathcal{N}(\boldsymbol{x}_{t-1}; \mu(\boldsymbol{x}_t, {t}), \sigma_t^2 \mathbf{I}),
    \label{eq:3}
    \end{aligned}
\end{equation}
where $\mu(\boldsymbol{x}_t, {t})$ is the mean of the Gaussian distribution, formulated as $\mu(\boldsymbol{x}_t, {t}) = \frac{\sqrt{{\alpha}_t} \left(1 - \bar{\alpha}_{t-1}\right)}{1 - \bar{\alpha}_t} \boldsymbol{x}_t + \frac{\sqrt{\bar{\alpha}_{t-1}} \left(1 - \alpha_t\right)}{1 - \bar{\alpha}_t} \widehat{\boldsymbol{x}}_0$, and $\sigma_t^2$ is the variance of the Gaussian distribution, formulated as $\sigma_t^2 = \dfrac{1 - \bar{\alpha}_{t-1}}{1 - \bar{\alpha}_t} \beta_t$.

\begin{figure*}[t!]
  \centering
  \footnotesize
  \begin{overpic}[width=0.9\textwidth]{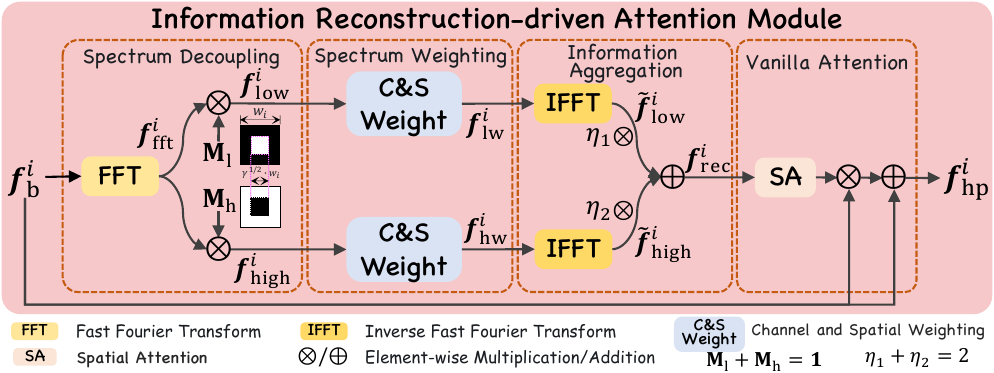}
  \end{overpic}
  \caption{ Illustration of the Information Reconstruction-driven Attention Module.
    }
  \label{fig:IRAM}
\end{figure*}

To achieve this goal of the reverse denoising process, a network is designed to predict $\widehat{\boldsymbol{x}}_0$ from $\boldsymbol{x}_t$.
Here, for convenience, in subsequent expressions, we use $\widehat{\boldsymbol{x}}_t$ instead of $\widehat{\boldsymbol{x}}_0$.
$\mu(\boldsymbol{x}_t, {t})$ is updated as follows:
\begin{equation}
   \begin{aligned}
      \mu(\boldsymbol{x}_t, {t}) = \frac{\sqrt{{\alpha}_t} \left(1 - \bar{\alpha}_{t-1}\right)}{1 - \bar{\alpha}_t} \boldsymbol{x}_t + \frac{\sqrt{\bar{\alpha}_{t-1}} \left(1 - \alpha_t\right)}{1 - \bar{\alpha}_t} \widehat{\boldsymbol{x}}_t.
    \label{eq:4}
    \end{aligned}
\end{equation}
%
Specifically, we achieve the network in our IPDiff.
Since our IPDiff is conditioned on the input ORSI, we formulate it as follows:
\begin{equation}
   \begin{aligned}
    \widehat{\boldsymbol{x}}_0/\widehat{\boldsymbol{x}}_t = {\psi}_{\theta}(\boldsymbol{x}_{t},t,\boldsymbol{I}),
    \label{eq:5}
    \end{aligned}
\end{equation}
where ${\psi}(\cdot)$ is our IPDiff, $\theta$ is its parameters, and $\boldsymbol{I}$ represents the input ORSI.
Thus, starting from the random Gaussian noise $\boldsymbol{x}_T \sim \mathcal{N}(\boldsymbol{0},\mathbf{I})$, our IPDiff can generate a set of $\widehat{\boldsymbol{x}}_t$ to be sampled for generating corresponding $\boldsymbol{x}_{t-1}$, achieving progressive denoising of $\boldsymbol{x}_T$ to recover to the clear $\boldsymbol{x}_0$.

\subsection{Network Overview}
\label{sec:Overview}

As shown in Fig.~\ref{fig:Framework}, the proposed IPDiff (\ie ${\psi}_{\theta}(\boldsymbol{x}_{t},t,\boldsymbol{I})$) consists of a prior network and a denoising network.
As their name suggests, the prior network is responsible for prior extraction, while the denoising network is responsible for noisy mask denoising.

Concretely, in the prior network, we adopt PVT-v2~\citep{2022PVTv2} as the backbone, whose inputs are the ORSI $\boldsymbol{I} \in \mathbb{R}^{3\!\times\!352\!\times\!352}$ and the noisy mask $\boldsymbol{x}_t\in \mathbb{R}^{1\!\times\!352\!\times\!352}$.
In the four blocks of PVT-v2, each block is embedded with the time token $\boldsymbol{t}$ of an appropriate size.
The time token $\boldsymbol{t}$ is derived from the timestep $t$.
After each block, we use two convolutional layers to adjust the channel number of its output features to 256.
Thus, we get four-level basic features denoted as $\{\boldsymbol{f}^{i}_{\rm b} \in \mathbb{R}^{c\!\times\!h_i\!\times\!w_i}\}^4_{i=1}$, where $c = 256$ and $h_i/w_i=\frac{352}{2^{i+1}}$.
$\boldsymbol{f}^{i}_{\rm b}$ is sent to IRAM to reconstruct information in the spectral domain to extract informative content, generating a conditional prior $\boldsymbol{f}^{i}_{\rm hp} \in \mathbb{R}^{c\!\times\!h_i\!\times\!w_i}$.
And four-level conditional priors form the hierarchical priors $\{\boldsymbol{f}^{i}_{\rm hp} \}^4_{i=1}$.
Moreover, we extract the saliency prior $\boldsymbol{S}_{t} \in [0,1]^{1\!\times\!h_1\!\times\!w_1}$ ($h_1/w_1=88$) from the integration of the hierarchical priors through a saliency prediction head, \ie SalPreHead.

Then, the saliency prior $\boldsymbol{S}_{t}$ and the hierarchical priors $\{\boldsymbol{f}^{i}_{\rm hp} \}^4_{i=1}$ are injected into the denoising network.
The input of the denoising network is the noisy mask $\boldsymbol{x}_t$, and the time token $\boldsymbol{t}$ is only embedded in the first encoder block.
$\boldsymbol{S}_{t}$ is integrated into the initial stage of the encoder, while $\{\boldsymbol{f}^{i}_{\rm hp} \}^4_{i=1}$ are hierarchically integrated into the encoder.
Four IPMs are embedded into the encoder to improve the feature representation ability, and they only appear in the training phase.
The output features of the encoder are transferred to the decoder through the information transmission.
Finally, $\widehat{\boldsymbol{x}}_t \in [0,1]^{1\!\times\!352\!\times\!352}$ is predicted through a SalPreHead.
According to Eq.~\ref{eq:4}, $\boldsymbol{x}_{t-1}$ is sampled from $\widehat{\boldsymbol{x}}_t$, and proceeds the next iteration of optimization.
After completing all $T$ iterations, we adopt $\widehat{\boldsymbol{x}}_0 $ as the final saliency map $\boldsymbol{S}_{\rm final} \in [0,1]^{1\!\times\!352\!\times\!352}$.



\subsection{Information Reconstruction-driven Attention Module}
\label{sec:D-SWSAM}

As is well known, objects in ORSIs have large size differences and varied shooting angles~\citep{2019LVNet,2021DAFNet,2022MJRBM}, which may lead to significant differences in the spatial representation of the same type of objects.
This brings great challenges to ORSI-SOD.
In addition, ORSIs often contain a large amount of spatial redundant information due to their large-scale coverage, such as large areas of uniform vegetation, water bodies, and sandy land~\citep{2019LVNet,2021DAFNet,2022MJRBM}.
This may lead to excessive attention to these redundant regions when processing ORSIs in the spatial domain, resulting in a waste of computing resources.
To alleviate the above issues, we propose the Information Reconstruction-driven Attention Module to reconstruct information in the spectral domain.

In the spectral domain, size differences essentially correspond to spectral differences, \ie large-size objects correspond to low-frequency components, while small-size objects correspond to high-frequency components.
Shooting angle variation can be described by the rotation or translation characteristics of the spectrum. 
Thus, in spectral domain reconstruction, the robustness of the model's perception to size and shooting angle variation can be enhanced through adaptive weighting of spectral components, such as adjusting the weights of low and high frequencies.
In addition, in the spectral domain, redundant information mostly corresponds to low-frequency components with concentrated energy, while key information (such as salient regions and edges) corresponds to specific high-frequency components. 
Through spectral domain reconstruction, the model can specifically focus on the spectral components containing key information, reduce the ineffective learning of redundant low frequencies, and improve the efficiency of model and the sensitivity to salient regions.
Therefore, our IRAM reconstructs information through adaptive spectrum decoupling and adaptive information aggregation, so as to extract useful content in a flexible and learnable manner.
We illustrate the detailed structure of IRAM in Fig.~\ref{fig:IRAM}.
In the following, we introduce IRAM in detail from four parts, \ie spectrum decoupling, spectrum weighting, information aggregation, and vanilla attention.

\textit{1) Spectrum Decoupling:}
The input of our IRAM is $\boldsymbol{f}^{i}_{\rm b} \in \mathbb{R}^{c\!\times\!h_i\!\times\!w_i}$.
We transform $\boldsymbol{f}^{i}_{\rm b}$ into the spectral domain, generating $\boldsymbol{f}^{i}_{\rm fft} \in \mathbb{R}^{c\!\times\!h_i\!\times\!w_i}$ as follows:
\begin{equation}
   \begin{aligned}
    \boldsymbol{f}^{i}_{\rm fft} = {\rm FFT}(\boldsymbol{f}^{i}_{\rm b} ),
    \label{eq:lowfre}
    \end{aligned}
\end{equation}
where ${\rm FFT}(\cdot)$ is the fast Fourier transform.
Then, we decouple $\boldsymbol{f}^{i}_{\rm fft}$ into low-frequency components and high-frequency components.

Concretely, we adopt two binary masks $ \{\mathbf{M}_{\rm l}, \mathbf{M}_{\rm h}\}\in \{0,1\}^{1 \times h_i\!\times\!w_i}$ to achieve spectrum decoupling.
As the spectrum decoupling part shown in Fig.~\ref{fig:IRAM}, we set the central square area of $\mathbf{M}_{\rm l}$ to 1, and the other areas to 0.
The hyperparameter $\gamma \in (0,1)$ controls the size of the central square area.
We formulate $\mathbf{M}_{\rm l}$ as follows: 
\begin{equation}
   \begin{aligned}
    \mathbf{M}^{h,w}_{\rm l}(\gamma) = 
    \begin{cases} 
    1, & h \in [(1-{\gamma}^{1/2})\frac{h_i}{2}, (1+{\gamma}^{1/2})\frac{h_i}{2}] \\
    &~~~\&w \in [(1-{\gamma}^{1/2})\frac{w_i}{2}, (1+{\gamma}^{1/2})\frac{w_i}{2}],\\
    0, & \text{otherwise}.
    \end{cases}
    \label{eq:lowfremask}
    \end{aligned}
\end{equation}
Following this, we can obtain $\mathbf{M}_{\rm h}$ as follows:
\begin{equation}
   \begin{aligned}
    \mathbf{M}_{\rm h} =  \mathbf{1}-\mathbf{M}_{\rm l},
    \label{eq:highfre}
    \end{aligned}
\end{equation}
where $\mathbf{1} \in \{1\}^{1\times h_i \times w_i}$.
These two binary masks are multiplied to $\boldsymbol{f}^{i}_{\rm fft}$ respectively to achieve spectrum decoupling, generating low-frequency components $\boldsymbol{f}^{i}_{\rm low} \in \mathbb{R}^{c\!\times\!h_i\!\times\!w_i}$ and high-frequency components $\boldsymbol{f}^{i}_{\rm high} \in \mathbb{R}^{c\!\times\!h_i\!\times\!w_i}$.

In practice, it is difficult to achieve an optimal manual decoupling of low-frequency components and high-frequency components for ORSI-SOD.
Therefore, we adopt an adaptive way to decouple the spectrum, that is, we set the hyperparameter $\gamma$ as a learnable hyperparameter.
In this way, our adaptive spectral decoupling can learn from ORSI data the appropriate spectral decoupling hyperparameter that adapts to complex ORSI scenes.

\textit{2) Spectrum Weighting:}
Since different types of information, such as background, shape, objects, and edge, are distributed in specific frequency bands of low-frequency components and high-frequency components~\citep{LowHighFre}, we continue to weight the obtained low-frequency components and high-frequency components.
Here, we perform the classical channel attention and spatial attention~\citep{2018CBAM} on $\boldsymbol{f}^{i}_{\rm low}$ and $\boldsymbol{f}^{i}_{\rm high}$, respectively, to achieve spectrum weighting, generating $\boldsymbol{f}^{i}_{\rm lw} \in \mathbb{R}^{c\!\times\!h_i\!\times\!w_i}$ and $\boldsymbol{f}^{i}_{\rm hw} \in \mathbb{R}^{c\!\times\!h_i\!\times\!w_i}$.
We formulate the above spectrum weighting as follows:
\begin{equation}
   \begin{aligned}
    \boldsymbol{f}^{i}_{\rm lw} &=  {\rm SA}( {\rm CA}(\boldsymbol{f}^{i}_{\rm low})), \\
    \boldsymbol{f}^{i}_{\rm hw} &=  {\rm SA}( {\rm CA}(\boldsymbol{f}^{i}_{\rm high})), 
    \label{eq:casa}
    \end{aligned}
\end{equation}
where ${\rm CA}(\cdot)$ and ${\rm SA}(\cdot)$ are channel attention and spatial attention, respectively.

Enhancing low-frequency and high-frequency components in both channel and spatial dimensions is conducive to highlighting objects in specific frequency bands.
This is a simple yet effective method.

\textit{3) Information Aggregation:}
Here, we perform the information aggregation on the weighted low-frequency and high-frequency components to achieve the information reconstruction.
We convert $\boldsymbol{f}^{i}_{\rm lw}$ and $\boldsymbol{f}^{i}_{\rm hw}$ back to the spatial domain, generating $\boldsymbol{\tilde{f}}^{i}_{\rm low} \in \mathbb{R}^{c\!\times\!h_i\!\times\!w_i}$ and $\boldsymbol{\tilde{f}}^{i}_{\rm high} \in \mathbb{R}^{c\!\times\!h_i\!\times\!w_i}$ as follows:
\begin{equation}
   \begin{aligned}
    \boldsymbol{\tilde{f}}^{i}_{\rm low}  &= {\rm IFFT}(\boldsymbol{f}^{i}_{\rm lw} ),\\
    \boldsymbol{\tilde{f}}^{i}_{\rm high} &= {\rm IFFT}(\boldsymbol{f}^{i}_{\rm hw} ),
    \label{eq:ifft}
    \end{aligned}
\end{equation}
where ${\rm IFFT}(\cdot)$ is the inverse fast Fourier transform.
Compared to the original $\boldsymbol{{f}}^{i}_{\rm low}$ and $\boldsymbol{{f}}^{i}_{\rm high}$, $\boldsymbol{\tilde{f}}^{i}_{\rm low}$ and $\boldsymbol{\tilde{f}}^{i}_{\rm high}$ have stronger representation capabilities.

At this time, we choose an adaptive way to aggregate $\boldsymbol{\tilde{f}}^{i}_{\rm low}$ and $\boldsymbol{\tilde{f}}^{i}_{\rm high}$ instead of simply aggregating them, which helps reconstruct information that is conducive to ORSI-SOD.
We set a learnable hyperparameter $\eta_1$ as the weight of $\boldsymbol{\tilde{f}}^{i}_{\rm low}$, and another learnable hyperparameter $\eta_2$ as the weight of $\boldsymbol{\tilde{f}}^{i}_{\rm high}$, reconstructing $\boldsymbol{{f}}^{i}_{\rm rec} \in \mathbb{R}^{c\!\times\!h_i\!\times\!w_i}$ as follows:
\begin{equation}
   \begin{aligned}
    \boldsymbol{{f}}^{i}_{\rm rec} = \eta_1 \otimes \boldsymbol{\tilde{f}}^{i}_{\rm low} + \eta_2 \otimes \boldsymbol{\tilde{f}}^{i}_{\rm high},
    \label{eq:InfAgg}
    \end{aligned}
\end{equation}
where $\{\eta_1,\eta_1\}\in (0,2)$ and $\eta_1 + \eta_2 =2$, and $\otimes$ is the element-wise multiplication.
Adaptive information aggregation is a soft aggregation that is more suitable for deep learning-based models than hard (or fixed) information aggregation (\ie $\eta_1 = \eta_2 =1$).
In this way, the reconstructed $\boldsymbol{{f}}^{i}_{\rm rec}$ is discriminative.

\textit{4) Vanilla Attention:}
Finally, based on the vanilla spatial attention~\citep{2018CBAM}, we adopt the reconstructed $\boldsymbol{{f}}^{i}_{\rm rec}$ to enhance the original input $\boldsymbol{{f}}^{i}_{\rm b}$, generating the output of IRAM $\boldsymbol{{f}}^{i}_{\rm hp} \in \mathbb{R}^{c\!\times\!h_i\!\times\!w_i}$ as follows:
\begin{equation}
   \begin{aligned}
    \boldsymbol{{f}}^{i}_{\rm hp} = ({\rm SA}(\boldsymbol{{f}}^{i}_{\rm rec}) \otimes \boldsymbol{{f}}^{i}_{\rm b}) \oplus \boldsymbol{{f}}^{i}_{\rm b},
    \label{eq:SA}
    \end{aligned}
\end{equation}
where $\oplus$ is the element-wise addition.

In summary, the adaptive spectrum decoupling and adaptive information aggregation endow IRAM with the ability to extract useful information from the basic features.
Its output $\boldsymbol{{f}}^{i}_{\rm hp}$ is also referred to as the conditional prior.
As shown in Fig.~\ref{fig:Framework}, by using IRAMs to process $\{\boldsymbol{f}^{i}_{\rm b}\}^4_{i=1}$, we obtain four-level conditional priors, forming the hierarchical priors $\{\boldsymbol{f}^{i}_{\rm hp}\}^4_{i=1}$.
The saliency prior originates from the hierarchical priors $\{\boldsymbol{f}^{i}_{\rm hp}\}^4_{i=1}$.
Therefore, IRAM serves as the cornerstone of multiple priors and plays a vital role in the prior network.


\begin{figure}[t!]
  \centering
  \begin{overpic}[width=1\columnwidth]{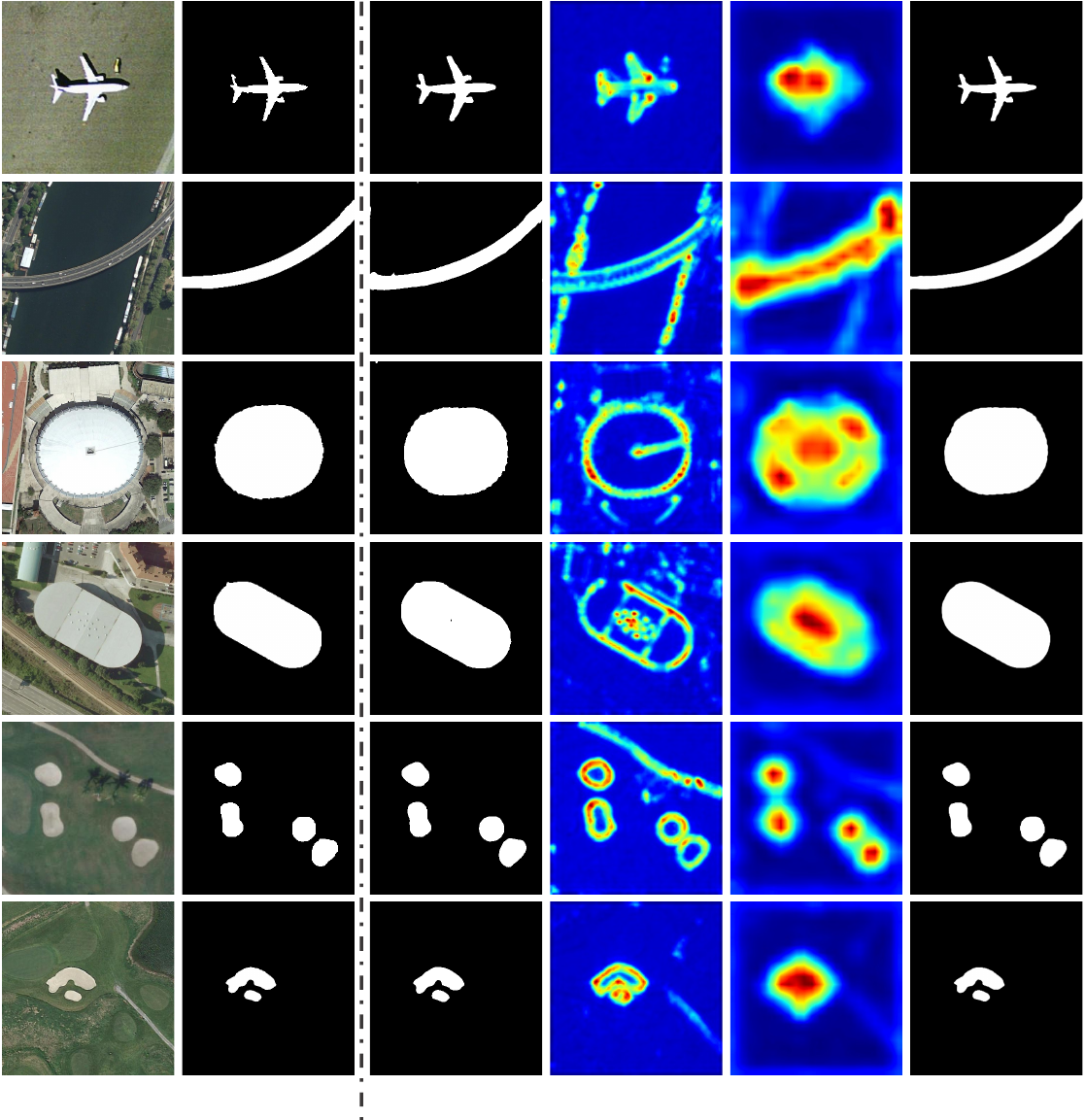}

    \put(2., 0){ ORSI }
    \put(20., 0){ GT }
    \put(32.8, 1.1){ Saliency  }
    \put(35.2,-2.2){ Prior }
    \put(53., 0.2){ $\boldsymbol{f}^{1}_{\rm hp}$ }
    \put(68.6, 0.2){ $\boldsymbol{f}^{3}_{\rm hp}$ }
    \put(83.3, 0.2){ $\boldsymbol{S}_{\rm final}$ }

  \end{overpic}
  \caption{Visualization of the saliency prior and the hierarchical priors ($\boldsymbol{f}^{1}_{\rm hp}$ and $\boldsymbol{f}^{3}_{\rm hp}$).}
  \label{fig:MultiPriorVisual}
\end{figure}



\subsection{Multi-Prior Guidance Denoising Network}
\label{sec:KTM} 
%
The denoising network is the core of the reverse denoising process.
To improve its effectiveness, we inject it with specially extracted conditional priors from ORSIs, \ie the saliency prior and the hierarchical priors.
In Fig.~\ref{fig:MultiPriorVisual}, we intuitively visualize the saliency prior and the hierarchical priors ($\boldsymbol{f}^{1}_{\rm hp}$ and $\boldsymbol{f}^{3}_{\rm hp}$).
We observe that the saliency prior indeed provides accurate positional guidance, which helps to localize the main object regions during the reverse denoising process.
For hierarchical priors, $\boldsymbol{f}^{1}_{\rm hp}$ is the shallow-level prior, exhibiting clear and rich details and textural information (\eg object edges and fine structures).
In contrast, the deep-level prior $\boldsymbol{f}^{3}_{\rm hp}$ presents strong semantic information, with responses concentrated on the object regions while effectively suppressing background interference.
We name our denoising network the multi-prior guidance denoising network.
The detailed structure of the multi-prior guidance denoising network is illustrated at the bottom of Fig.~\ref{fig:Framework}.

The multi-prior guidance denoising network is built on the encoder-decoder architecture.
Its main input is the noisy mask $\boldsymbol{x}_t$ accompanied by the time token $\boldsymbol{t}$.
We first adopt an encoder block (\ie EncBlk1\footnote{EncBlk1 consists of a 7$\times$7 convolutional layer, a ResNet block, and a 3$\times$3 convolutional layer.}) to extract basic features $\boldsymbol{f}^{1}_{\rm en} \in \mathbb{R}^{c\!\times\!h_1\!\times\!w_1}$ from $\boldsymbol{x}_t$.
Since the saliency prior $\boldsymbol{S}_t$ is a coarse saliency map, it contains positional information of salient objects.
We adopt a saliency encoder block (\ie SalEncBlk\footnote{SalEncBlk consists of a 3$\times$3 convolutional layer.}) to extract such positional feature $\boldsymbol{f}_{\rm sal} \in \mathbb{R}^{c\!\times\!h_1\!\times\!w_1}$ from $\boldsymbol{S}_t$, and fuse it with $\boldsymbol{f}^{1}_{\rm en}$ at the initial stage of the encoder.
Our approach of injecting the saliency prior to the denoising network is unique, and it helps to stabilize the position of objects in complex ORSI scenes.
Since $\boldsymbol{x}_t$ and $\boldsymbol{S}_t$ are essentially grayscale images, the information that can be extracted is relatively limited.
Therefore, we also incorporate $\boldsymbol{f}^{1}_{\rm hp}$, which can provide detail information, into them to achieve feature fusion through a fusion block (\ie FuBlk\footnote{FuBlk consists of a 3$\times$3 convolutional layer.}), generating $\boldsymbol{{f}}^{1}_{\rm fu} \in \mathbb{R}^{c\!\times\!h_1\!\times\!w_1}$.
Then, we arrange three EncBlks\footnote{EncBlk2, EncBlk3, and EncBlk4 each consist of a 3$\times$3 convolutional layer.} to continuously extract basic features of different levels in the encoder, getting $\{\boldsymbol{f}^{i}_{\rm en} \in \mathbb{R}^{c\!\times\!h_i\!\times\!w_i}\}^4_{i=2}$.
In this process, $\{\boldsymbol{f}^{i}_{\rm hp}\}^4_{i=2}$ are hierarchically injected into the encoder, providing specific detail and semantic information of objects.

\begin{figure}[t!]
  \centering
  \footnotesize
  \begin{overpic}[width=0.95\columnwidth]{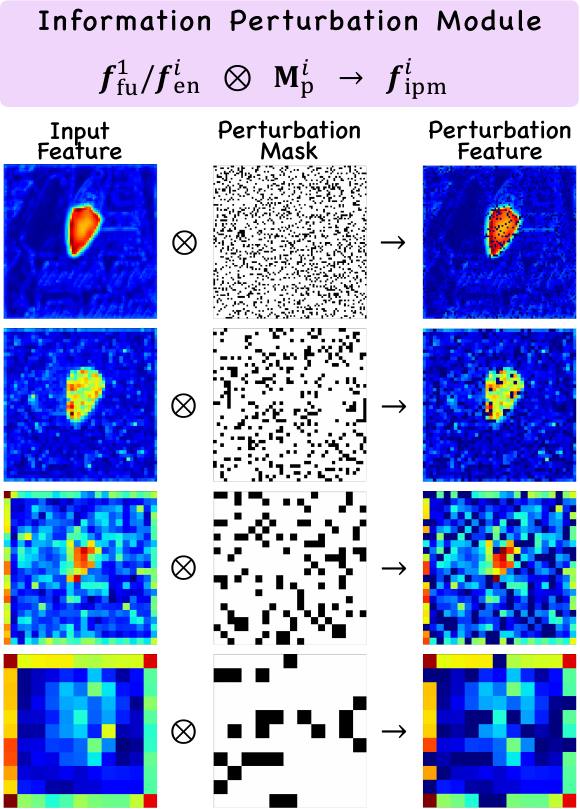}
  \end{overpic}
  \caption{ Visualization of operations and features in the Information Perturbation Module.
From top to bottom are features in IPM$^1$ to IPM$^4$, respectively.    }
  \label{fig:IPM}
\end{figure}

The stability of feature extraction is undoubtedly important for denoising a noisy mask.
We introduce IPM into the above encoder of the multi-prior guidance denoising network, and arrange it after FuBlk, EncBlk2, EncBlk3, and EncBlk4, generating the corresponding perturbation feature $\boldsymbol{{f}}^{i}_{\rm ipm} \in \mathbb{R}^{c\!\times\!h_i\!\times\!w_i}$.
We formulate IPM as follows:
\begin{equation}
   \begin{aligned}
    \boldsymbol{{f}}^{i}_{\rm ipm} = 
    \begin{cases} 
    \boldsymbol{{f}}^{i}_{\rm fu} \otimes \mathbf{M}^{i}_{\rm p}, & i=1,\\
    \boldsymbol{{f}}^{i}_{\rm en} \otimes \mathbf{M}^{i}_{\rm p}, & i=2,3,4,
    \end{cases}
    \label{eq:IPM}
    \end{aligned}
\end{equation}
where $\mathbf{M}^{i}_{\rm p}$ is the perturbation mask, belonging to $\{0,1\}^{c\!\times\!h_i\!\times\!w_i}$, and the proportion of 0 (\ie perturbation rate) in $\mathbf{M}^{i}_{\rm p}$ is $r \in [0,1] $.
We visualize features in IPM with $r =20\% $ in Fig.~\ref{fig:IPM}.
The perturbation mask masks part of the information in the input features through multiplication.
This operation enables the encoder to adaptively recover the lost information during the feature extraction process, thereby naturally improving the feature representation and anti-interference capabilities of the encoder.
Notably, IPMs are only embedded into the encoder in the training phase, and will not disturb the testing phase.


The decoder corresponds to the encoder.
It contains four blocks (\ie DecBlk\footnote{DecBlk1 is composed of a sequence of a 3$\times$3 convolutional layer, a 2$\times$ upsampling layer, a 3$\times$3 convolutional layer, another 2$\times$ upsampling layer, and a final 3$\times$3 convolutional layer. DecBlk2, DecBlk3, and DecBlk4 each consist of two 3$\times$3 convolutional layers followed by a 2$\times$ upsampling layer.}).
Since the priors containing position, detail, and semantic information are hierarchically injected into the encoder, we introduce an information transmission to hierarchically transfer them to the decoder to fully utilize this information.
With the help of SalPreHead, our denoising network outputs the optimized $\widehat{\boldsymbol{x}}_t$.
$\widehat{\boldsymbol{x}}_t$ is used to sample $\boldsymbol{x}_{t-1}$ for the next iterative optimization until $\widehat{\boldsymbol{x}}_0$ is obtained as the final saliency map $\boldsymbol{S}_{\rm final}$.
As shown in the last column of Fig.~\ref{fig:MultiPriorVisual}, benefiting from the positional guidance of the saliency prior and the multi-scale detail and semantic cues provided by the hierarchical priors, the final saliency map $\boldsymbol{S}_{\rm final}$ exhibits high consistency with the ground truth.


In summary, our multi-prior guidance denoising network can achieve stable denoising with the assistance of multiple priors and IPMs.
Specifically, the operation of IPM is relatively simple, but its improvement in feature representation and anti-interference capabilities of our denoising network is significant.
With the collaboration of all components, our denoising network can effectively resist the noise and interference specific to optical imaging (such as illumination changes, atmospheric scattering, and cloud occlusion) and has good adaptability to complex ORSI scenes.

\subsection{Spatial-Spectral Collaborative Alignment-based Hybrid Loss Function}
\label{sec:HybridLossFunction}
%
As illustrated in Fig.~\ref{fig:Framework}, our IPdiff has two items that need to be supervised in the training phase, \ie $\widehat{\boldsymbol{x}}_t \in [0,1]^{1\!\times\!352\!\times\!352}$ and $\boldsymbol{S}_{t} \in [0,1]^{1\!\times\!88\!\times\!88}$.
Different previous methods that only supervise the network in the spatial domain~\citep{2019LVNet,2025MRBINet,2025DPUFormer}, we simultaneously align $\widehat{\boldsymbol{x}}_t$ and $\boldsymbol{S}_{t}$ with GTs in both spatial and spectral domains to improve the efficiency of network training.
Accordingly, we construct a spatial-spectral collaborative alignment-based hybrid loss function ${L}_{\rm spa\&spe}$ as follows:
\begin{equation}
   \begin{aligned}
    {L}_{\rm spa\&spe}  = \underbrace{{L}_{\rm spa-base} + {L}_{\rm spa-edge}}_{{\rm Spatial}} + \underbrace{{L}_{\rm spe}}_{{\rm Spectral}}.
    \label{eq:spa&spe}
    \end{aligned}
\end{equation}
%

For the spatial item, we not only retain the traditional ${L}_{\rm spa-base}$ including the weighted binary cross-entropy loss (${\ell}_{\rm wbce}$) and the weighted intersection-over-union loss (${\ell}_{\rm wiou}$), but also introduce the edge loss ${L}_{\rm spa-edge}$ to directly focus on edges of salient objects through the binary cross-entropy loss (${\ell}_{\rm bce}$).
We formulate ${L}_{\rm spa-base}$ and ${L}_{\rm spa-edge}$ as follows:
\begin{equation}
   \begin{aligned}
    {L}_{\rm spa-base}  = {\ell}_{\rm wbce} (\boldsymbol{S}, \boldsymbol{G}) + {\ell}_{\rm wiou} (\boldsymbol{S}, \boldsymbol{G}),
    \label{eq:base}
    \end{aligned}
\end{equation}
\begin{equation}
   \begin{aligned}
    {L}_{\rm spa-edge} = {\ell}_{\rm bce}(|\boldsymbol{S}-{\rm AP}(\boldsymbol{S})|,|\boldsymbol{G}-{\rm AP}(\boldsymbol{G})|) ,
    \label{eq:edge}
    \end{aligned}
\end{equation}
where $\boldsymbol{S}$ is predicted saliency map, $\boldsymbol{G}$ is GT, and 
${\rm AP}(\cdot)$ is a 3$\times$3 average pooling layer with stride of 1 and padding of 1.

For the spectral item ${L}_{\rm spe}$, we separate the real and imaginary parts of the predicted saliency map in the spectral domain, and adopt the structural similarity index loss (${\ell}_{\rm ssim}$) to align these two parts with their corresponding parts of GT, respectively.
We formulate ${L}_{\rm spe}$ as follows:
\begin{equation}
   \begin{aligned}
    {L}_{\rm spe} = & \big[ \underbrace{{\ell}_{\rm ssim}\big({\rm Re}({\rm FFT}(\boldsymbol{S})),{\rm Re}({\rm FFT}(\boldsymbol{G}))\big)}_{{\rm Real }} +\\ 
   &  \underbrace{{\ell}_{\rm ssim}\big({\rm Im}({\rm FFT}(\boldsymbol{S})),{\rm Im}({\rm FFT}(\boldsymbol{G}))\big)}_{{\rm Imaginary}} \big] /2,
    \label{eq:spe}
    \end{aligned}
\end{equation}
where ${\rm Re}(\cdot)$ and ${\rm Im}(\cdot)$ mean the real part and the imaginary part, respectively.
${L}_{\rm spe}$ provides a new perspective for network training, which can increase the diversity of supervision on the traditional spatial supervision.

We adopt the hybrid loss function ${L}_{\rm spa\&spe}$ to align $\widehat{\boldsymbol{x}}_t$ and $\boldsymbol{S}_{t}$ with GTs, and formulate the total loss function ${L}_{\rm total}$ as follows:
\begin{equation}
   \begin{aligned}
    {L}_{\rm total}  = {L}_{\rm spa\&spe} (\widehat{\boldsymbol{x}}_t, \boldsymbol{G}) + 0.5 \cdot {L}_{\rm spa\&spe} ( {\rm Up}(\boldsymbol{S}_t), \boldsymbol{G}) ,
    \label{eq:TotalLoss}
    \end{aligned}
\end{equation}
where $ {\rm Up}(\cdot)$ is the upsampling operation, and $\boldsymbol{G}$ belongs to $\{0,1\}^{1\!\times\!352\!\times\!352}$. 
Notably, we set the coefficient of $\boldsymbol{S}_{t}$ to 0.5, which is smaller than that of $\widehat{\boldsymbol{x}}_t$.
This is because $\boldsymbol{S}_{t}$ with a small size tends to have a larger loss value, and a smaller coefficient can make the training focus more on the final output of IPDiff $\widehat{\boldsymbol{x}}_t$.

\begin{table*}[t!]
  \centering
  \renewcommand{\arraystretch}{1.25}
  \renewcommand{\tabcolsep}{0.8mm}
  \caption{
   Quantitative and model complexity comparisons with state-of-the-art relevant methods on EORSSD, ORSSD, and ORSI-4199 datasets.
   $\uparrow$ indicates that the larger the better, while $\downarrow$ the opposite.
   We mark the best result in \textbf{bold} and the second best result in \textit{\underline{italic}}.
   }
\label{table:QuantitativeResults}
  
\resizebox{1\textwidth}{!}{
\begin{tabular}{r|c|c|ccc|cccc|cccc|cccc}
\midrule[1pt]    
 \multirow{2}{*}{\normalsize{Methods}}
 & \multirow{2}{*}{\normalsize{Type}}
 & Input
 & Param
 & FLOPs
 & Speed
 & \multicolumn{4}{c|}{EORSSD~\citep{2021DAFNet}} 
 & \multicolumn{4}{c|}{ORSSD~\citep{2019LVNet}} 
 & \multicolumn{4}{c}{ORSI-4199~\citep{2022MJRBM}}  \\
 
 \cline{7-10} \cline{11-13} \cline{14-18}
       &
       &  Size
       &(M)$\downarrow$
       &(G)$\downarrow$
       &(fps)$\uparrow$
       & $S_{\alpha}\uparrow$ & $F_{\beta}^{\rm{max}}\uparrow$ & $E_{\xi}^{\rm{max}}\uparrow$ & $ \mathcal{M}\downarrow$
   	          & $S_{\alpha}\uparrow$ & $F_{\beta}^{\rm{max}}\uparrow$ & $E_{\xi}^{\rm{max}}\uparrow$ & $ \mathcal{M}\downarrow$
	          & $S_{\alpha}\uparrow$ & $F_{\beta}^{\rm{max}}\uparrow$ & $E_{\xi}^{\rm{max}}\uparrow$ & $ \mathcal{M}\downarrow$\\
	     
\midrule[1pt]

R3Net$_{18}$~\citep{2018R3Net}   	& CN &  300$^2$&  56.1&  47.5&  2&.8184 & .7498 & .9483 & .0171
									 & .8141 & .7456 & .8913 &  .0399
									 & .8142 & .7847  & .8880  & .0401 \\
PoolNet$_{19}$~\citep{2019PoolNet}  & CN &  \tiny{400$\times$300}&  53.6&  123.4&  25& .8207 & .7545  & .9292  & .0210
									    & .8403 & .7706  & .9343 & .0358 
									    & .8271 & .8010  & .8964  & .0541 \\
EGNet$_{19}$~\citep{2019EGNet}  	& CN &  \tiny{380$\times$320} &  108.0&  291.9&  9& .8601 & .7880  & .9570  & .0110  
									 & .8721 & .8332  & .9731 & .0216  
									 & .8464 & .8267  & .9161  & .0440  \\								    
GCPA$_{20}$~\citep{2020GCPA}  	& CN &  320$^2$&  67.1&  54.3&  23& .8869 & .8347  & .9524  & .0102  
									 & .9026 & .8687  & .9509  & .0168  
									 & - & - & - & -  \\ 
MINet$_{20}$~\citep{2020MINet}  	& CN &  320$^2$&  47.5&  146.3&  12& .9040 & .8344  & .9442  & .0093
									   & .9040 & .8761  & .9545  & .0144  
									   & - & - & - & -  \\ 
ITSD$_{20}$~\citep{2020ITSD}  		& CN &  228$^2$ &  17.1&  54.5&  16& .9050 & .8523  & .9556  & .0106
									   & .9050 & .8735  & .9601 & .0165  
									   & - & - & - & -  \\ 
GateNet$_{20}$~\citep{2020GateNet} & CN &  384$^2$&  100.0&  108.3&  25& .9114 & .8566  & .9610 & .0095
									    & .9186 & .8871 & .9664 & .0137  
									    & - & - & - & -  \\ 
CSNet$_{20}$~\citep{2020CSNet} 	& CN &  224$^2$ &  0.14&  0.7&  38& .8364 & .8341  & .9535  & .0169
									   & .8910 & .8790  & .9628  & .0186 
									   & .8241 & .8124  & .9096  & .0524 \\
SAMNet$_{21}$~\citep{2021SAMNet} 	& CN &  336$^2$ &  1.33&  0.5 &  44 & .8622 & .7813 & .9421 & .0132
									   & .8761 & .8137  & .9478  & .0217 
									   & .8409 & .8249  & .9186  & .0432 \\	
HVPNet$_{21}$~\citep{2021HVPNet} 	& CN &  336$^2$ &  1.23&  1.1&  26 & .8734 & .8036  & .9482  & .0110
									   & .8610 & .7938  & .9320  & .0225 
									   & .8471 & .8295  & .9201  & .0419 \\								   
SUCA$_{21}$~\citep{2021SUCA}  	& CN &  256$^2$&  117.7&  56.4&  24& .8988 & .8229 & .9520  & .0097
									   & .8989 & .8484  & .9584  & .0145
									   & .8794 & .8692  & .9438  & .0304 \\
PA-KRN$_{21}$~\citep{2021PAKRN}  & CN &  600$^2$&  141.1&  617.7&  16& .9192 & .8639  & .9616  & .0104
									   & .9239 & .8890  & .9680  & .0139 
									   & .8491 & .8415 & .9280  & .0382 \\									   
VST$_{21}$~\citep{2021VST}        & TN &  224$^2$&  44.1&  23.2&  23& .9208 & .8716  & .9743  & .0067
									   & .9365 & .9095  & .9810  & .0094 
									   & .8790 & .8717  & .9481  & .0281 \\
DPORTNet$_{22}$~\citep{2022DPORTNet}  &  CN &  352$^2$&  18.9&  60.4&  16& .8960 & .8363  & .9423  & .0150 
									    & .8827 & .8309 & .9214  & .0220 
									    & .8094 & .7789  & .8759 & .0569 \\
DNTD$_{22}$~\citep{2022DNTD}  &  CN &  224$^2$&  28.8&  8.1&  -& .8957 & .8189  & .9378  & .0113 
									    & .8698 & .8231  & .9286  & .0217 
									    & .8444 & .8310  & .9158  & .0425 \\
ICON$_{23}$~\citep{2023ICON}  &  TN  &  352$^2$&  65.7&  61.8&  34& .9185 & .8622  & .9687  & .0073 
									    & .9256 & .8939  & .9704  & .0116  
									    & .8752 & .8763  & .9521  & .0282 \\
\hline
LVNet$_{19}$~\citep{2019LVNet}  	  & CO &  128$^2$ &  207.0 &  - &   1 & .8630 & .7794  & .9254  & .0146 
									      & .8815 & .8263 & .9456 & .0207
									      & - & - & - & -  \\ 
DAFNet$_{21}$~\citep{2021DAFNet}    & CO &  128$^2$ &  29.3 &  68.5 &  26 & .9166 & .8614 &  \textbf{.9861} & .0060
									      & .9191 & .8928 & .9771  & .0113 
									      & - & - & - & -  \\ 
SARNet$_{21}$~\citep{2021SARNet} & CO &  336$^2$ &  25.9 &  129.7  &  47 & .9240 & .8719  & .9620 & .0099
									   & .9134 & .8850  & .9557  & .0187 
									   & - & - & - & - \\ 							     
MJRBM$_{22}$~\citep{2022MJRBM} & CO &  352$^2$  &  43.5 &  95.7 &  32& .9197 & .8656  & .9646 & .0099
									   & .9204 & .8842  & .9623  & .0163  
									   & .8593 & .8493  & .9311  & .0374 \\
EMFINet$_{22}$~\citep{2022EMFINet} & CO &  256$^2$ &  107.3 &  480.9 &  25 & .9290 & .8720  & .9711  & .0084
									     & .9366 & .9002  & .9737  & .0109  
									     & .8675 & .8584  & .9340  & .0330  \\
CorrNet$_{22}$~\citep{2022CorrNet} 		& CO &  256$^2$ &  4.09 &  21.1 &  100 & .9289 & .8778  & .9696  & .0083
									   &  .9380 &  .9129  & .9790  & .0098  
									   & .8623 & .8560  & .9330  & .0366  \\
MCCNet$_{22}$~\citep{2022MCCNet} 	  & CO &  256$^2$ &  67.6 &  112.8 &  95 & .9327 & .8904  & .9755  & .0066
				       			& .9437 & .9155 & .9800 & .0087 
									  & .8746 & .8690 & .9413 & .0316 \\			       		  
HFANet$_{22}$~\citep{2022HFANet}  &  HO &  448$^2$ &  60.5 &  68.3 &  26 &  .9380 & .8876  & .9740 & .0070 
									    & .9399 & .9112  & .9770  & .0092 
									    & .8767 & .8700  & .9431  & .0314 \\ 
ERPNet$_{23}$~\citep{2022ERPNet}  & CO &  224$^2$ &  56.5 &  87.2 &  50  & .9210 & .8632  & .9603 & .0089 
									   & .9254  & .8974 & .9710  & .0135  
									   & .8670 & .8553  & .9290 & .0357 \\
SeaNet$_{23}$~\citep{2023SeaNet}  & CO &  288$^2$ &  2.76 &  1.7 &  96 & .9208 & .8649  & .9710 & .0073 
									   & .9260  & .8942 & .9767  & .0105  
									   & .8772 & .8653  & .9426 & .0308 \\
ACCoNet$_{23}$~\citep{2023ACCoNet} 	  & CO &  256$^2$ &  102.5 &  179.9 &  81 & .9290 & .8837  & .9727  & .0074
									   & .9437 &  .9149 & .9796  & .0088 
									   & .8775 & .8686  & .9412 & .0314 \\
GeleNet$_{23}$~\citep{2023GeleNet}				 & TO &  352$^2$ &  25.4 &  11.7 &  30 & .9376 & .8923 & \textit{\underline{.9828}}  & .0064 
									   & .9469 & .9254 & .9860  &  .0079  
									   & .8862 & .8842  & .9544 & .0264  \\	

GLGCNet$_{23}$~\citep{2023GLGCNet}			& TO &  352$^2$ &  25.1 &  9.8 &  21  & .9375 & .8924 & .9803  & .0055 
									   & .9488 & .9236 & .9864 &  .0071  
									   & .8839 & .8808 & .9508 & .0274  \\	

MIRGNet$_{24}$~\citep{2024MIRGNet}  &  CO &  256$^2$ &  78.7 &  136.2 &  46 & .9383 & \textit{\underline{.8930}} & .9789 & .0056
										& .9455 & .9192  & .9812  & .0081
									   	& - & - & - & -  \\ 
SAFINet$_{24}$~\citep{2024SAFINet}  &  CO &  288$^2$ &  3.12 &  7.6 &  - & .9267 & .8799  & .9732  & .0065
										& .9401 & .9106  & .9786  & .0086
									   	& - & - & - & -  \\ 
TSCNet$_{24}$~\citep{2024TSCNet}  &  CO &  256$^2$ &  103.6 &  116.8 & 59 & .9376 & .8900 & .9765  & .0061 
									    & .9428 & .9198  & .9850  & .0081 
									    & .8783 & .8771  & .9486  & .0295 \\ 

RAGRNet$_{24}$~\citep{2024RAGRNet}  & CO &  256$^2$ &  35.6 &  17.8 &  31 & .9361 & .8852 & .9785 & .0057
										& \textit{\underline{.9507}} & .9242 & .9861 & .0066
										& .8811 & .8811 & .9492 & .0284 \\ 

SFANet$_{24}$~\citep{2024SFANet}  &  CO &  256$^2$ &  25.1 &  7.7 &  - & .9349 & .8833  & .9769 & .0058
										& .9453 & .9192 & .9830 & .0077
										& .8761 & .8710 & .9447 & .0292 \\ 

UDCNet-R$_{24}$~\citep{2024UDCNet}  &  CO &  352$^2$&  72.3&  101.2&  43& .9310 & .8821 & .9774 & .0056 
										& .9497 & .9239 & .9850 & .0068
										& .8802 & .8808 & .9515 & .0266 \\ 
										
TLCKDNet$_{24}$~\citep{2024TLCKDNet}  &  TO &  256$^2$ &  50.0 &  31.7 &  32 & .9350 & .8843 & .9788 & .0056
										& .9421 & .9114 & .9794 & .0082
									   	& - & - & - & - \\ 
PRNet$_{24}$~\citep{2024PRNet}  & TO &  352$^2$ &  20.8 &  8.5 &  21 & .9276 & .8684 & .9784 & \textit{\underline{.0054}}
										& .9459 & .9177 & .9848 & .0075
										& .8873 & .8819 & .9527 & .0272 \\ 
ADSTNet$_{24}$~\citep{2024ADSTNet}  &  HO &  256$^2$ &  62.1 &  27.7 &  40 & .9311 & .8804 & .9769 & .0065
										& .9379 & .9124  & .9807  & .0086
									   	& .8710 & .8698  & .9433  & .0318 \\ 
	
SOLNet$_{25}$~\citep{2025SOLNet}  &  CO &  256$^2$ &  6.52 &  8.1 &  161 & .9171 & .8609 & .9623 & .0078
										& .9284 & .9012 & .9734 & .0111
									   	& - & - & - & - \\ 
SggNet$_{25}$~\citep{2025SggNet}  &  CO &  288$^2$ &  2.70 &  1.4 &  108 & .9278 & .8871 & .9762  & .0068
										& .9342 & .9030 & .9758 & .0111
									   	& - & - & - & - \\ 	
BCARNet$_{25}$~\citep{2025BCARNet}  &  CO &  352$^2$ &  24.0 &  7.0 &  - & .9361 & .8871 & .9761 & \textit{\underline{.0054}}
										& .9465 & .9196 & .9833 & .0071
										& .8757 & .8689 & .9407 & .0306 \\ 
MRBINet$_{25}$~\citep{2025MRBINet}  &  CO &  256$^2$ &  32.4 &  42.8 &  9 & .9351 & .8852 & .9766 & .0056 
										& .9474 & .9199 & .9851 & .0069 
										& .8824 & .8800 & .9489 & .0268  \\ 

DPU-Former$_{25}$~\citep{2025DPUFormer}  &  TO &  352$^2$ &  44.2 &  32.5 &  43 & \textit{\underline{.9401}} & \textit{\underline{.8930}} & .9816 & .0056
										& .9412 & \textit{\underline{.9263}} & \textit{\underline{.9868}} & \textit{\underline{.0062}}
										& .8833 & \textit{\underline{.8877}} & \textit{\underline{.9547}} &  .0263 \\ 

\hline								   				
EnsemDiff$_{22}$~\citep{2022EnsemDiff}  &  Diff &  224$^2$ &  124.0 &  190.2&  0.03& .8516 & .7690 & .9063 & .0173 
									     & .8795 & .8189 & .9343 & .0245
									    & .8051 & .7583  & .8681 & .0558 \\ 
MedSegDiffv2$_{24}$~\citep{2024MedSegDiffv2}  &  Diff &  256$^2$ &  139.7 &  260.6 &  0.03& .7429 & .5739  & .7918  & .0342
									    & .7727 & .6457  & .8225  & .0508
									    & .7490 & .6835  & .8202  & .0754 \\ 				    
CamoDiff$_{25}$~\citep{2025CamoDiffusion}  &  Diff &  384$^2$ &  71.7 &  60.3 &  8 & .9314 & .8890  & .9801  & .0113 
									    & .9343 & .9215 & .9863  & .0187
									    & \textit{\underline{.8891}} & .8804 & .9539 & \textit{\underline{.0250}} \\ 
							   
\hline
\hline
                                        
\textbf{IPDiff (Ours)}				 & Diff &  352$^2$ &  82.6&  70.4 &  4 & \textbf{.9461} & \textbf{.9033}  & \textbf{.9861} & \textbf{.0044} 
									    & \textbf{.9557} & \textbf{.9362}  & \textbf{.9915}  & \textbf{.0054}
									    & \textbf{.8907} & \textbf{.8885} & \textbf{.9572} & \textbf{.0237} \\ 	
\bottomrule[1pt]
\multicolumn{14}{l}{\footnotesize{CN/TN: CNN-/Transformer-based NSI-SOD method, CO/TO: CNN-/Transformer-based ORSI-SOD method.}} \\
\multicolumn{14}{l}{\footnotesize{HO: Hybrid backbone-based ORSI-SOD method, Diff: Diffusion-based method.}} \\
\multicolumn{14}{l}{\footnotesize{We reserve two decimal places for the parameter count of lightweight methods.}} \\
\end{tabular}
} 
\end{table*}

\section{Experiments}
\label{sec:exp}

\subsection{Experimental Setup}
\label{sec:ExpProtocol}
\textit{1) Datasets:}
We conduct experiments on three commonly used ORSI-SOD datasets, \ie ORSSD~\citep{2019LVNet}, EORSSD~\citep{2021DAFNet}, and ORSI-4199~\citep{2022MJRBM} datasets.
The ORSSD dataset\footnote{https://li-chongyi.github.io/proj\_optical\_saliency.html} is a small dataset, containing 800 ORSIs and GTs, among which 600 images form the training set and 200 images form the test set.
The EORSSD dataset\footnote{https://github.com/rmcong/DAFNet\_TIP20} extends the ORSSD dataset to 2,000 ORSIs and GTs, among which 1,400 images form the training set and 600 images form the test set.
The ORSI-4199 dataset\footnote{https://github.com/wchao1213/ORSI-SOD} is a big dataset, containing 4,199 ORSIs and GTs, among which 2,000 images form the training set and 2,199 images form the test set.
Following the traditional mode of ORSI-SOD~\citep{2023SeaNet,2023GeleNet,2024SFANet}, we train and test our IPDiff separately on each of the three datasets.

\textit{2) Implementation Details:}
We conduct experiments using PyTorch framework and an NVIDIA RTX 3090 GPU.
The input size of our IPDiff is set to 352$\times$352.
We initialize the backbone (\ie PVT-v2~\citep{2022PVTv2}) of our prior network with the pre-trained parameters.
We rotate and flip inputs for data augmentation.
We adopt the AdamW optimizer to conduct network training for 150 epochs with an initial learning rate of $1e^{-4}$ and a batch size of 16.
The perturbation rate $r$ of IPM is set to $20\%$.
For the denoising diffusion process, we set the total timestep $T$ of our IPDiff to 10.

\textit{3) Evaluation Metrics:}
We use four quantitative evaluation metrics from an evaluation tool\footnote{https://github.com/MathLee/MatlabEvaluationTools} to assess the performance of our IPDiff and all compared methods on ORSSD, EORSSD, and ORSI-4199 datasets, including S-measure ($S_{\alpha}$, $\alpha$ = 0.5)~\citep{Fan2017Smeasure},
maximum F-measure ($F_{\beta}^{\rm{max}}$, $\beta^{2}$ = 0.3)~\citep{Fmeasure},
maximum E-measure ($E_{\xi}^{\rm{max}}$)~\citep{Fan2018Emeasure}, and
mean absolute error (MAE, $\mathcal{M}$).
The first three metrics are better when they are larger, while the last one is better when it is smaller.

In addition, we adopt the parameter count, the computational cost, and the inference speed (without I/O time) to evaluate the model complexity.
The first two metrics are better when they are smaller, while the last one is better when it is larger.

\subsection{Comparison with State-of-the-arts}
We compare our IPDiff with 46 state-of-the-art NSI-SOD methods, ORSI-SOD methods, and diffusion-driven segmentation methods on the EORSSD, ORSSD, and ORSI-4199 datasets.
NSI-SOD methods include R3Net~\citep{2018R3Net}, PoolNet~\citep{2019PoolNet}, EGNet~\citep{2019EGNet}, GCPA~\citep{2020GCPA}, MINet~\citep{2020MINet}, ITSD~\citep{2020ITSD}, GateNet~\citep{2020GateNet}, CSNet~\citep{2020CSNet}, SAMNet~\citep{2021SAMNet}, HVPNet~\citep{2021HVPNet}, SUCA~\citep{2021SUCA}, PA-KRN~\citep{2021PAKRN}, VST~\citep{2021VST}, DPORTNet~\citep{2022DPORTNet}, DNTD~\citep{2022DNTD}, and ICON~\citep{2023ICON}.
ORSI-SOD methods include LVNet~\citep{2019LVNet}, DAFNet~\citep{2021DAFNet}, SARNet~\citep{2021SARNet}, MJRBM~\citep{2022MJRBM}, EMFINet~\citep{2022EMFINet}, CorrNet~\citep{2022CorrNet}, MCCNet~\citep{2022MCCNet}, HFANet~\citep{2022HFANet}, ERPNet~\citep{2022ERPNet}, SeaNet~\citep{2023SeaNet}, ACCoNet~\citep{2023ACCoNet}, GeleNet~\citep{2023GeleNet}, GLGCNet~\citep{2023GLGCNet}, MIRGNet~\citep{2024MIRGNet}, SAFINet~\citep{2024SAFINet}, TSCNet~\citep{2024TSCNet}, RAGRNet~\citep{2024RAGRNet}, SFANet~\citep{2024SFANet}, UDCNet-R~\citep{2024UDCNet}, TLCKDNet~\citep{2024TLCKDNet}, PRNet~\citep{2024PRNet}, ADSTNet~\citep{2024ADSTNet}, SOLNet~\citep{2025SOLNet}, SggNet~\citep{2025SggNet}, BCARNet~\citep{2025BCARNet}, MRBINet~\citep{2025MRBINet}, and DPU-Former~\citep{2025DPUFormer}.
Diffusion-driven segmentation methods include EnsemDiff~\citep{2022EnsemDiff}, MedSegDiffv2~\citep{2024MedSegDiffv2}, and CamoDiff~\citep{2025CamoDiffusion}.
The saliency maps for the above methods are obtained from the authors or by running public source codes.

\begin{figure*}[t!]
    \centering
    \scriptsize
	\begin{overpic}[width=1\textwidth]{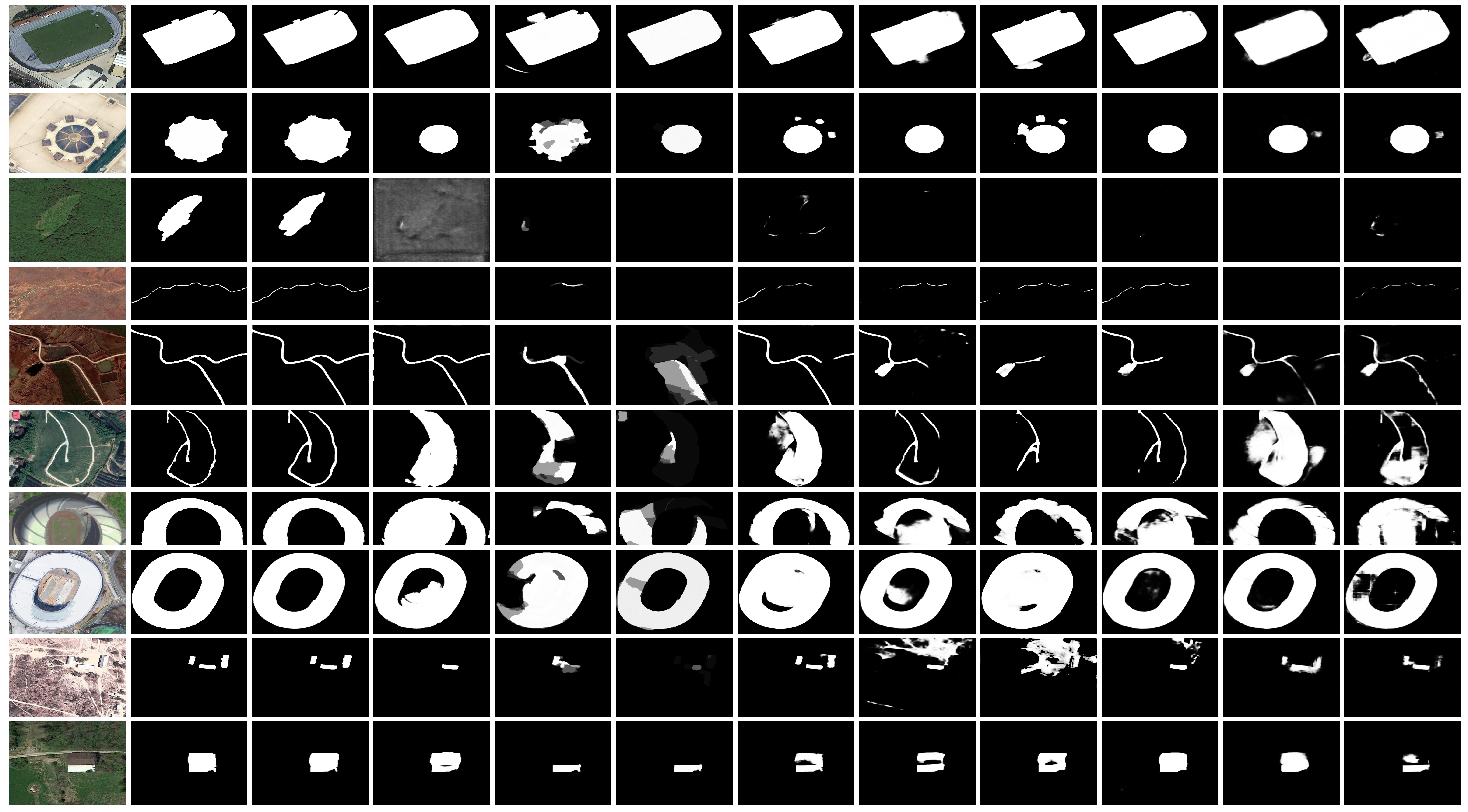}
    \put(-1.,49.75){ \begin{sideways}{\tiny{ORSSD}}\end{sideways} }
    \put(-1.,44.75){ \begin{sideways}{\tiny{4199}}\end{sideways} }
    \put(-1.,38.35){ \begin{sideways}{\tiny{ORSSD}}\end{sideways} }
    \put(-1.,32.55){ \begin{sideways}{\tiny{EORSSD}}\end{sideways} }
    \put(-1.,28.85){ \begin{sideways}{\tiny{4199}}\end{sideways} }
    \put(-1.,23.20){ \begin{sideways}{\tiny{4199}}\end{sideways} }
    \put(-1.,18.55){ \begin{sideways}{\tiny{4199}}\end{sideways} }
    \put(-1.,13.60){ \begin{sideways}{\tiny{4199}}\end{sideways} }
    \put(-1.,7.55){ \begin{sideways}{\tiny{4199}}\end{sideways} }
    \put(-1.,1.05){ \begin{sideways}{\tiny{ORSSD}}\end{sideways} }

    \put(2.0,-1.2){ ORSI }
    \put(11.05,-1.2){ GT}
    \put(18.55,-1.2){ \textbf{Ours}}
    \put(24.9,-1.2){ CamoDiff  }
    \put(32.05,-1.2){ MedSegDiffv2 }
    \put(42.05,-1.2){ EnsemDiff }
    \put(49.6,-1.2){ DPU-Former }
    \put(58.9,-1.2){ MRBINet }
    \put(66.52,-1.2){ BCARNet }
    \put(74.82,-1.2){ RAGRNet }
    \put(84.6,-1.2){ ICON } 
    \put(93.37,-1.2){ VST } 
    
    \end{overpic}
	\caption{Qualitative comparisons with nine representative state-of-the-art methods.
    The source dataset is shown on the left side of the ORSI. We abbreviate the ORSI-4199 dataset as 4199.
    }
    \label{fig:VisualComparisons}
\end{figure*}
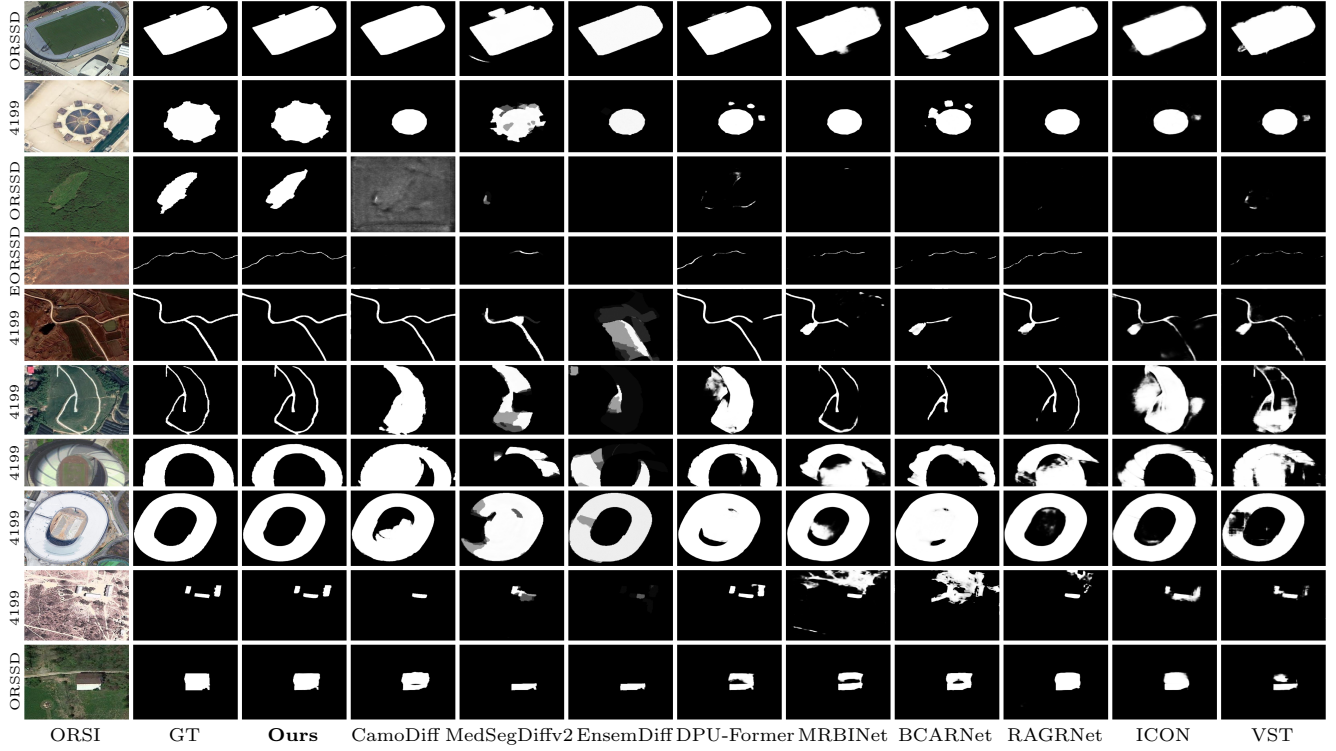

\textit{1) Quantitative and Model Complexity Comparison:}
In Tab.~\ref{table:QuantitativeResults}, we report the quantitative and model complexity comparison results of our IPDiff and other 46 compared methods on the EORSSD, ORSSD, and ORSI-4199 datasets.
Overall, our method outperforms all compared methods in all four evaluation metrics.
On the EORSSD dataset, our $S_{\alpha}$ is one of the only two methods that exceed 0.94, one of which is 0.9401 for DPU-Former~\citep{2025DPUFormer} and the other is 0.9461 for our IPDiff.
Our $F_{\beta}^{\rm{max}}$ is the only one that exceeds 0.90, reaching 0.9033, which is 1.03\% higher than 0.8930 of the second-highest MIRGNet~\citep{2024MIRGNet} and DPU-Former~\citep{2025DPUFormer}.
Our $\mathcal{M}$ is the only one below 0.005, reaching 0.0044, which is 0.001 lower than 0.0054 of the second-lowest PRNet~\citep{2024PRNet} and BCARNet~\citep{2025BCARNet}.
On the ORSSD dataset, our $S_{\alpha}$ is one of the only two methods that exceed 0.95, one of which is 0.9507 for RAGRNet~\citep{2024RAGRNet} and the other is 0.9557 for our IPDiff.
Our $F_{\beta}^{\rm{max}}$ is the only one that exceeds 0.93, reaching 0.9362, which is 0.99\% higher than 0.9263 of the second-highest DPU-Former~\citep{2025DPUFormer}.
Our $E_{\xi}^{\rm{max}}$ is the only one that exceeds 0.99, reaching an amazing 0.9915.
Our $\mathcal{M}$ is the only one below 0.006, reaching 0.0054.
On the ORSI-4199 dataset, our $S_{\alpha}$ is the only one that exceeds 0.89, reaching 0.8907.
Moreover, compared to NSI-SOD methods, specialized ORSI-SOD methods show more excellent performance.
Diffusion-driven segmentation methods perform on par with some specialized ORSI-SOD methods, achieving impressive performance.
But they are inferior to our specifically designed IPDiff for ORSI-SOD.

Specifically, we also provide a detailed analysis of our dynamic optimization strategy and different representative paradigms.
Most existing ORSI-SOD methods, such as GeleNet and DPU-Former, are typical one-shot baselines.
These methods map an input to the saliency map in a single forward pass.
As shown in Tab.~\ref{table:QuantitativeResults}, although DPU-Former and GeleNet achieve competitive results, our IPDiff outperforms them across all metrics, demonstrating the necessity of an optimization strategy.
Notably, their advantage lies in efficiency.
PRNet and GLGCNet are multi-stage refinement baselines, which progressively generate saliency maps within the decoder and perform gradual refinement.
However, unlike our IPDiff which treats optimization as a dynamic denoising process, these methods rely on static architectural stages.
The experimental results show that IPDiff significantly surpasses PRNet and GLGCNet. 
This indicates that the diffusion-style dynamic optimization strategy is inherently more effective than conventional multi-stage refinement.

\begin{table*}[!t]
\centering
\caption{Ablation study on the effectiveness of each component in IRAM. 
InfRec, SpeDec, SpeWei, and InfAgg are the abbreviations for information reconstruction, spectrum decoupling, spectrum weighting, and information aggregation, respectively.
The best result of each metric is shown in \textbf{bold}.
  }
\label{Ablation_IRAM}
\renewcommand{\arraystretch}{1.25}
\renewcommand{\tabcolsep}{1.2mm}
\begin{tabular}{c|cc|cc|cc|cc||cccc}
\toprule

  \multirow{2}{*}{No.} & \multirow{2}{*}{SA} & \multirow{2}{*}{Direct InfRec}  & \multicolumn{2}{c|}{SpeDec} & \multicolumn{2}{c|}{SpeWei} & \multicolumn{2}{c||}{InfAgg} 
 & \multicolumn{4}{c}{EORSSD~\citep{2021DAFNet}} \\
 
 \cline{4-5} \cline{6-7}  \cline{8-9} \cline{10-13} 
    & & & Fix ${\gamma}$ & Adp ${\gamma}$  & LowFre  & HighFre & Fix $\{\eta_1,\eta_2\}$  & Adp $\{\eta_1,\eta_2\}$
    &$S_{\alpha}\uparrow$ & $F_{\beta}^{\rm{max}}\uparrow$ 
    & $E_{\xi}^{\rm{max}}\uparrow$  & $ \mathcal{M}\downarrow$ \\
\midrule
1 &  \Checkmark &                      &                      &                      &                      &                      &                      &                      
& .9375  & .8952  & .9793   & .0072  \\ 
2 &  \Checkmark & \Checkmark  &                      &                      &                      &                      &                      &                      
& .9448  & .9004  & .9822   & \textbf{.0040}  \\ 
\hline
3 &  \Checkmark &                       & \Checkmark  &   &  \Checkmark   & &\Checkmark &                      
& .9434  & .8998  & .9827   & .0046  \\ 
4 &  \Checkmark &                     &  \Checkmark  &  &   & \Checkmark & \Checkmark    &          
& .9444  & .8993  & .9811   & .0043  \\ 
5 &  \Checkmark &                       &  \Checkmark  &  & \Checkmark & \Checkmark & \Checkmark   &         
& .9453  & .9010  & .9828   & .0047  \\ 
\hline
6 &  \Checkmark &                       & \Checkmark &  & \Checkmark & \Checkmark &  &  \Checkmark 
& .9450  & .9019  & .9839   & .0046  \\ 
7 &  \Checkmark &                       &      & \Checkmark & \Checkmark & \Checkmark  & \Checkmark  & 
& .9458  & .9007  & .9831   & .0045  \\ 

\hline
\hline
8 &  \Checkmark &                      &   & \Checkmark & \Checkmark & \Checkmark  & &  \Checkmark  
& \textbf{.9461}  & \textbf{.9033}  & \textbf{.9861}  & .0044 \\
\bottomrule
\multicolumn{13}{l}{\footnotesize{Fix means fixed hyperparameter, while Adp means adaptive hyperparameter.}} \\
\end{tabular}
\end{table*}

Regarding the model complexity, the parameter count and computational cost of our IPDiff are 82.6M and 70.4G FLOPs, respectively, achieving an inference speed of 4 fps with a 352$\times$352 input.
Obviously, our IPDiff has a significant disadvantage in inference speed.
This is because our IPDiff is built on a diffusion-based architecture, which has an inherent requirement of $T$ denoising steps in the reverse denoising process.
But it achieves a better balance between performance and efficiency as compared to two diffusion-based methods, \ie EnsemDiff and MedSegDiffv2.
Notably, although its iterative nature results in a higher computational cost than most one-shot methods, our IPDiff remains lighter than several heavy CNN-based competitors in parameter count, \eg EGNet, ACCoNet, and TSCNet.

\textit{2) Qualitative Comparison:}
%
%
In Fig.~\ref{fig:VisualComparisons}, we show the saliency maps generated from our IPDiff and nine representative state-of-the-art methods, including three diffusion-driven segmentation methods (CamoDiff, MedSegDiffv2, and EnsemDiff), four ORSI-SOD methods (DPU-Former, MRBINet, BCARNet, and RAGRNet), and two NSI-SOD methods (ICON and VST).
There are ten cases in Fig.~\ref{fig:VisualComparisons}, totaling six challenging ORSI scenes from three datasets.
The first scene is objects with fine structures (\ie 1st and 2nd cases).
Our IPDiff outlines fine structures, while other methods only locate the position of objects.
The second one is the low-contrast scene (\ie 3rd and 4th cases).
Most methods lose low-contrast objects, but our IPDiff accurately highlights them.
The third one is objects with complex topology (\ie 5th and 6th cases), with road being a typical representative.
The complex topology causes many methods to fail, but our IPDiff handles it well.
The fourth one is hollow objects (\ie 7th and 8th cases), with gymnasium being a typical representative.
Most methods fail to properly segment the hollow, but our method segments the gymnasium with a complete hollow.
The fifth one is the overexposed scene (\ie 9th case).
The special ORSI scene caused by optical imaging makes objects difficult to distinguish, resulting in objects in the saliency maps of many methods being stuck together.
The last one is objects composed of regions with significant appearance differences (\ie last case).
Significant appearance differences put some methods in a dilemma.

\subsection{Ablation Studies}
\label{Ablation Studies}
We conduct comprehensive ablation experiments to assess the effectiveness of each part of our IPDiff on the EORSSD dataset.
Specifically, we evaluate
1) the effectiveness of each component in IRAM,
2) the rationality of each component in the denoising network,
3) the influence of the perturbation rate $r$ in IPM,
4) the influence of the number of IPMs,
5) the importance of each component of the hybrid loss function,
6) the effectiveness of the dynamic optimization strategy,
7) the influence of the timestep $T$, and
8) the generalization ability evaluation.

\textit{1) Effectiveness of Each Component in IRAM:}
IRAM achieves effective information reconstruction in the spectral domain through spectrum decoupling, spectrum weighting, and information aggregation.
Specifically, we perform spectrum decoupling and information aggregation in an adaptive way with learnable hyperparameters.
Here, we conduct variants to gradually verify the effectiveness of each component in IRAM.
We report the quantitative ablation results in Tab.~\ref{Ablation_IRAM}.

\begin{table}[!t]
\centering
\caption{Ablation study on the rationality of each component in the denoising network. The best result of each metric is shown in \textbf{bold}.}
\label{Ablation_denoising}
\renewcommand{\arraystretch}{1.25}
\renewcommand{\tabcolsep}{1.9mm}
{%
\begin{tabular}{c||cccc}
\toprule
 \multirow{2}{*}{Models}   & \multicolumn{4}{c}{EORSSD~\citep{2021DAFNet}}   \\
 \cline{2-5}
  & $S_{\alpha}\uparrow$ & $F_{\beta}^{\rm{max}}\uparrow$ & $E_{\xi}^{\rm{max}}\uparrow$ & $ \mathcal{M}\downarrow$\\
\midrule
\textit{w/o SalPrior}  & .9414  & .8935  & .9788   & .0051  \\
\textit{w/ IntegratedPrior}  & .9407  & .8936  & .9823   & .0046  \\ 
\textit{w/o IPM}  & .9446  & .8994  & .9811   & .0050  \\
\textit{w/o InfTrans}  & .9396  & .8915  & .9800   & .0050  \\

\hline
\hline
\textbf{Ours}  & \textbf{.9461}  & \textbf{.9033}  & \textbf{.9861}  & \textbf{.0044}    \\ \bottomrule
\end{tabular}%
}
\end{table}

First, we verify the effectiveness of information reconstruction in the spectral domain.
As shown in No.1 of Tab.~\ref{Ablation_IRAM}, we only keep the vanilla attention and discard the three parts for information reconstruction.
The performance degradation is obvious, \ie 0.86\% in $S_{\alpha}$ and 0.81\% in $F_{\beta}^{\rm{max}}$, which proves the effectiveness of information reconstruction in the spectral domain.
Moreover, we verify the necessity of spectrum decoupling for information reconstruction.
As shown in No.2 of Tab.~\ref{Ablation_IRAM}, we directly perform spectrum weighting on the entire spectrum without decoupling low-frequency and high-frequency components, and then reconstruct information, \ie Direct InfRec.
Spectrum weighting improves performance, but is suboptimal due to the absence of spectrum decoupling.

Then, we verify the effectiveness of spectrum weighting.
As shown in No.3$\sim$5 of Tab.~\ref{Ablation_IRAM}, we decouple the spectrum and aggregate information in a fixed way, and provide three combinations of weighting low-frequency and high-frequency components.
Abandoning either high-frequency weighting or low-frequency weighting is inferior to weighting both low-frequency and high-frequency components.
This indicates that weighting both low-frequency and high-frequency components is more conducive to mining useful information for ORSI-SOD.

Last, we verify the effectiveness of adaptive hyperparameters for spectrum decoupling and information aggregation.
As shown in No.5$\sim$8 of Tab.~\ref{Ablation_IRAM}, we provide four combinations of ${\gamma}$ and $\{\eta_1,\eta_2\}$, \ie fixed ${\gamma}$ and fixed $\{\eta_1,\eta_2\}$, fixed ${\gamma}$ and adaptive $\{\eta_1,\eta_2\}$, adaptive ${\gamma}$ and fixed $\{\eta_1,\eta_2\}$, and adaptive ${\gamma}$ and adaptive $\{\eta_1,\eta_2\}$.
Adaptive hyperparameters obviously help our IPDiff better decouple the spectrum and reconstruct information that can adapt to the complex ORSI scenes.

\begin{table}[!t]
\centering
\caption{Ablation study on the influence of the perturbation rate $r$ in IPM. The best result of each metric is shown in \textbf{bold}.}
\label{Ablation_perturbationrate}
\renewcommand{\arraystretch}{1.25}
\renewcommand{\tabcolsep}{1.9mm}
{%
\begin{tabular}{c||cccc}
\toprule
 \multirow{2}{*}{Perturbation rate $r$}   & \multicolumn{4}{c}{EORSSD~\citep{2021DAFNet}}   \\
 \cline{2-5}
  & $S_{\alpha}\uparrow$ & $F_{\beta}^{\rm{max}}\uparrow$ & $E_{\xi}^{\rm{max}}\uparrow$ & $ \mathcal{M}\downarrow$\\
\midrule
$0\%$  & .9446  & .8994  & .9811   & .0050  \\
$10\%$  & .9428  & .9001  & .9846   & .0045  \\
$20\%$ \textbf{(Ours)}  & \textbf{.9461}  & \textbf{.9033}  & .9861  & \textbf{.0044}    \\ 
$30\%$ & .9453  & .9028  & \textbf{.9862}   & .0046  \\
$40\%$  & .9431  & .9022  & .9858   & .0045  \\
$50\%$  & .9438  & .8987  & .9824   & .0045  \\
$60\%$  & .9428  & .8974  & .9837   & \textbf{.0044}  \\
\bottomrule
\end{tabular}%
}
\end{table}

\textit{2) Rationality of Each Component in the Denoising Network:}
To investigate the individual contribution of components in our multi-prior guidance denoising network, we design four variants:
1) removing the saliency prior, \ie \textit{w/o SalPrior}, 
2) changing the injection way of hierarchical priors to integrated prior injection, \ie \textit{w/ IntegratedPrior},
3) removing IPM in the training phase, \ie \textit{w/o IPM}, and
4) removing information transmission to prohibit information from being transferred to the decoder hierarchically, \ie \textit{w/o InfTrans}.
We report the quantitative ablation results in Tab.~\ref{Ablation_denoising}.

Without the saliency prior to provide positional information, the performance of \textit{w/o SalPrior} degrades quite a lot, with a drop of nearly 1.0\% in $F_{\beta}^{\rm{max}}$.
\textit{w/ IntegratedPrior} is the same as CamoDiff~\citep{2025CamoDiffusion}, which first integrates hierarchical priors into a comprehensive prior and then injects it into the highest level of the encoder of the denoising network.
However, this injection approach is not as effective as hierarchically injecting priors into the encoder for extracting valid features.
\textit{w/o IPM} reduces the feature extraction and anti-interference capabilities of our denoising network, resulting in a 0.5\% decrease in $E_{\xi}^{\rm{max}}$.
Surprisingly, the performance degradation of \textit{w/o InfTrans} is quite severe (\ie 1.18\% in $F_{\beta}^{\rm{max}}$), which indicates that transferring the information extracted in the encoder to the decoder can improve the decoder's perception of salient objects.

\textit{3) Influence of the Perturbation Rate $r$ in IPM:}
The perturbation rate $r$ in IPM determines how much information is discarded.
We conduct ablation experiments to experimentally analyze its influence.
As shown in Tab.~\ref{Ablation_perturbationrate}, when the perturbation rate $r$ is 20\%, our IPDiff performs best, and is higher than the variant that does not discard information (\ie $r$ is 0\%).
We observe that as more and more information is discarded (with $r$ increasing from 20\% to 60\%), the performance deteriorates increasingly.
This is consistent with our intuitive understanding, that is, the more information is discarded, the more difficult it is to reconstruct features.
Therefore, the feature extraction and anti-interference capabilities will be weakened.

\begin{table}[!t]
\centering
\caption{Ablation study on the influence of the number of IPMs. The best result of each metric is shown in \textbf{bold}.}
\label{Ablation_numberIPMs}
\renewcommand{\arraystretch}{1.25}
\renewcommand{\tabcolsep}{1.9mm}
{%
\begin{tabular}{c||cccc}
\toprule
 \multirow{2}{*}{Number of IPMs}   & \multicolumn{4}{c}{EORSSD~\citep{2021DAFNet}}   \\
 \cline{2-5}
  & $S_{\alpha}\uparrow$ & $F_{\beta}^{\rm{max}}\uparrow$ & $E_{\xi}^{\rm{max}}\uparrow$ & $ \mathcal{M}\downarrow$\\
\midrule
0  & .9446  & .8994  & .9811   & .0050  \\
1  & .9432  & .8995  & .9819   & .0047  \\
2  & .9441  & .8984  & .9823   & .0047  \\
3  & .9454  & \textbf{.9033}  & .9848   & .0046  \\
4 \textbf{(Ours)}  & \textbf{.9461}  & \textbf{.9033}  & \textbf{.9861}  & \textbf{.0044}    \\ 
\bottomrule
\end{tabular}%
}
\end{table}

\begin{table}[!t]
\centering
\caption{Ablation study on the importance of each component of the hybrid loss function. The best result of each metric is shown in \textbf{bold}.}
\label{Ablation_loss}
\renewcommand{\arraystretch}{1.25}
\renewcommand{\tabcolsep}{0.5mm}
{%
\begin{tabular}{c||cccc}
\toprule
 \multirow{2}{*}{Losses}   & \multicolumn{4}{c}{EORSSD~\citep{2021DAFNet}}   \\
 \cline{2-5}
  & $S_{\alpha}\uparrow$ & $F_{\beta}^{\rm{max}}\uparrow$ & $E_{\xi}^{\rm{max}}\uparrow$ & $ \mathcal{M}\downarrow$\\
\midrule
${L}_{\mathrm{spa-base}}$          & .9395  & .8924  & .9803   & .0066     \\
${L}_{\mathrm{spa-base}}$+${L}_{\mathrm{spa-edge}}$ (${L}_{\mathrm{spa}}$)          & .9410  & .8964  & .9824   & \textbf{.0044}    \\
${L}_{\mathrm{spa-base}}$+${L}_{\mathrm{spe}}$         & .9416  & .8960  & .9821  & .0064     \\
\hline
\hline
${L}_{\mathrm{spa}}$+${L}_{\mathrm{spe}}$   (${L}_{\rm spa\&spe}$)      & \textbf{.9461}          & \textbf{.9033}          & \textbf{.9861}   & \textbf{.0044}       \\ 
\bottomrule
\end{tabular}%
}
\end{table}

\textit{4) Influence of the Number of IPMs:}
In the denoising network, we hierarchically integrate four IPMs into its encoder during the training phase to enhance its anti-interference capability.
To evaluate the influence of the number of IPMs on the anti-interference capability, we conduct variants with 0, 1, 2, and 3 IPMs.
We report the experimental results in Tab.~\ref{Ablation_numberIPMs}.
As observed, the performance generally increases monotonically with the number of IPMs.
However, the performance in $S_\alpha$ and $F_\beta^{\text{max}}$ exhibits slight fluctuations when the number of IPMs is low (1 or 2).
We believe this is because the IPM introduces a local feature masking (\ie local perturbation) mechanism to simulate noise and interference.
When only a limited number of IPMs are equipped, the local perturbation may exceed the reconstruction capacity of the denoising network at that specific stage, thereby causing a slight drop in performance. 
As the number of IPMs increases to 3 and 4, we observe a consistent improvement across all metrics.
This trend indicates a synergistic effect across the hierarchical stages of our denoising network.
By equipping all four encoder blocks with IPMs, the denoising network is forced to learn robust feature representations across multiple scales, ranging from fine-grained textures to high-level semantics.
Consequently, this configuration achieves peak performance with an $S_{\alpha}$ of 0.9461 and an $\mathcal{M}$ of 0.0044, endowing our denoising network with a comprehensive anti-interference capability.

\begin{table}[!t]
\centering
\caption{The performance of the $t$-th step saliency maps ($\widehat{\boldsymbol{x}}_{t}$). The best result of each metric is shown in \textbf{bold}.}
\label{Ablation_DynamicOoptimization}
\renewcommand{\arraystretch}{1.25}
\renewcommand{\tabcolsep}{2.1mm}
{%
\begin{tabular}{c||cccc}
\toprule
 \multirow{2}{*}{$t$-th step}   & \multicolumn{4}{c}{EORSSD~\citep{2021DAFNet}}   \\
 \cline{2-5}
  & $S_{\alpha}\uparrow$ & $F_{\beta}^{\rm{max}}\uparrow$ & $E_{\xi}^{\rm{max}}\uparrow$ & $ \mathcal{M}\downarrow$\\
\midrule
$\widehat{\boldsymbol{x}}_{9}$   & .9452  & .9020  & .9851   & .0050  \\
$\widehat{\boldsymbol{x}}_{8}$   & .9453  & .9023  & .9851   & .0049  \\
$\widehat{\boldsymbol{x}}_{7}$   & .9455  & .9025  & .9850   & .0049  \\
$\widehat{\boldsymbol{x}}_{6}$   & .9456  & .9026  & .9850   & .0048  \\
$\widehat{\boldsymbol{x}}_{5}$   & .9462  & .9027  & .9850   & .0048  \\
$\widehat{\boldsymbol{x}}_{4}$   & .9460  & .9026  & .9850   & .0048  \\
$\widehat{\boldsymbol{x}}_{3}$   & .9460  & .9029  & .9851   & .0047  \\
$\widehat{\boldsymbol{x}}_{2}$   & \textbf{.9464}  & .9032  & .9856   & .0046  \\
$\widehat{\boldsymbol{x}}_{1}$   & .9462  & .9032  & .9858   & .0045  \\
\hline
\hline
$\widehat{\boldsymbol{x}}_{0}$ \textbf{(Ours)}  & .9461  & \textbf{.9033}  & \textbf{.9861}  & \textbf{.0044}    \\ 
\bottomrule
\end{tabular}%
}
\end{table}

\begin{figure*}[t!]
    \centering
	\begin{overpic}[width=1\textwidth]{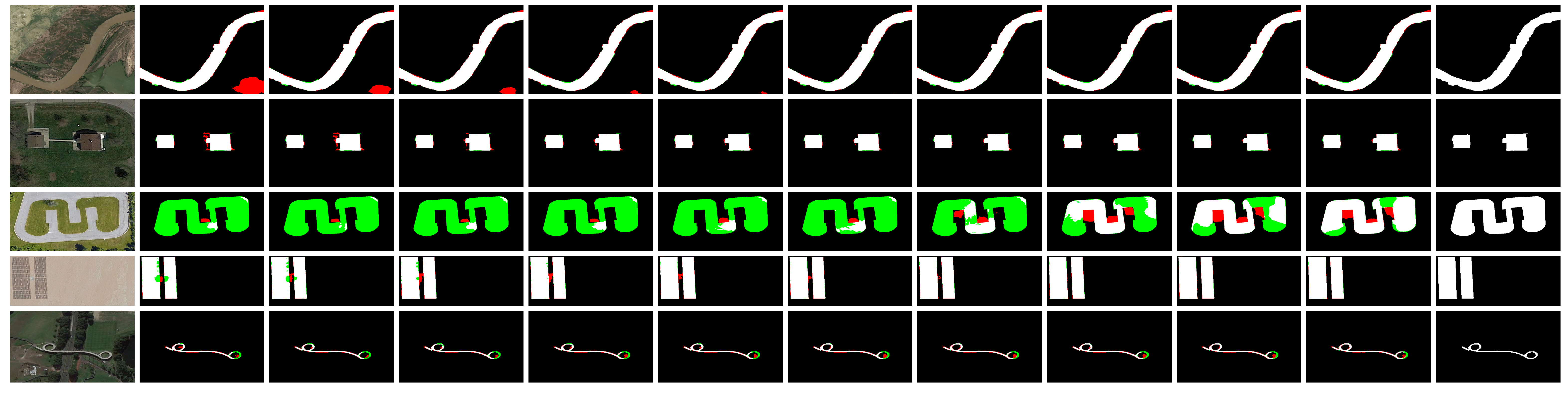}

    \put(1.7,-1.2){ ORSI }
    \put(11.45,-1.2){ $\widehat{\boldsymbol{x}}_{9}$}
    \put(19.75,-1.2){ $\widehat{\boldsymbol{x}}_{8}$}
    \put(27.95,-1.2){ $\widehat{\boldsymbol{x}}_{7}$  }
    \put(36.35,-1.2){ $\widehat{\boldsymbol{x}}_{6}$ }
    \put(44.45,-1.2){ $\widehat{\boldsymbol{x}}_{5}$ }
    \put(52.65,-1.2){ $\widehat{\boldsymbol{x}}_{4}$ }
    \put(61.05,-1.2){ $\widehat{\boldsymbol{x}}_{3}$ }
    \put(69.35,-1.2){ $\widehat{\boldsymbol{x}}_{2}$ }
    \put(77.65,-1.2){ $\widehat{\boldsymbol{x}}_{1}$ }
    \put(85.85,-1.2){ $\widehat{\boldsymbol{x}}_{0}$ } 
    \put(93.45,-1.2){ GT } 
    
    \end{overpic}%
	\caption{Visualization of the dynamic optimization from $\widehat{\boldsymbol{x}}_{9}$ to $\widehat{\boldsymbol{x}}_{0}$ in our IPDiff. In these saliency maps, we highlight the true positives in white, the true negatives in black, the false positives in red, and the false negatives in green. Please zoom in for details.
    }
    \label{fig:VisualDynamicOptimization}
\end{figure*}
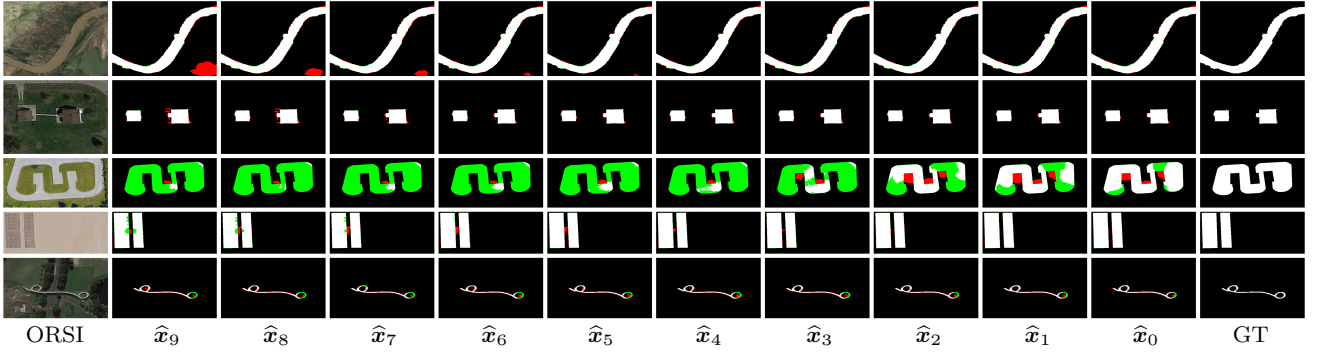

\textit{5) Importance of Each Component of the Hybrid Loss Function:}
As formulated in Eq.~\ref{eq:spa&spe}, our hybrid loss function ${L}_{\rm spa\&spe}$ consists of three components, including ${L}_{\rm spa-base}$, ${L}_{\rm spa-edge}$, and ${L}_{\rm spe}$, to supervise the predicted saliency map in both spatial and spectral domains.
To investigate the individual contribution of components in our ${L}_{\rm spa\&spe}$, we design three loss variants for network training:
1) only using the traditional ${L}_{\rm spa-base}$,
2) only supervising the predicted saliency map in the spatial domain, and
3) combining ${L}_{\rm spa-base}$ and ${L}_{\rm spe}$.
We report the quantitative ablation results in Tab.~\ref{Ablation_loss}.
The traditional ${L}_{\rm spa-base}$ is suboptimal.
With the help of edge loss ${L}_{\rm spa-edge}$, ${L}_{\rm spa}$ and ${L}_{\rm spa\&spe}$ both achieve performance improvements, especially with a reduction of 33.33\% and 31.25\% respectively in $\mathcal{M}$.
By additionally applying unique spectral domain alignment, three metrics (\ie $S_{\alpha}$, $F_{\beta}^{\rm{max}}$, and $E_{\xi}^{\rm{max}}$) significantly improve.
With all three losses training together, our IPDiff achieves the best performance.

\textit{6) Effectiveness of the Dynamic Optimization Strategy:}
The dynamic optimization strategy is the core of our IPDiff.
To verify the effectiveness of this strategy, we report the performance of saliency maps across all $T$ ($T$=10) steps, \ie $\{\widehat{\boldsymbol{x}}_{t}\}_{t=0}^{9}$, in Tab.~\ref{Ablation_DynamicOoptimization}.
We observe that $F_{\beta}^{\text{max}}$ exhibits a consistent upward trend, while $\mathcal{M}$ continuously decreases, indicating a steady improvement in model performance across the dynamic optimization process. 
For $S_\alpha$, although slight fluctuations are observed, it generally follows an increasing trajectory, peaking at $\widehat{\boldsymbol{x}}_{2}$ (0.9464) and maintaining a superior value at $\widehat{\boldsymbol{x}}_{0}$ (0.9461) compared to the initial $\widehat{\boldsymbol{x}}_{9}$ (0.9452). 
$E_{\xi}^{\text{max}}$ remains stable in the early stages, and shows a marked increase during the final steps.
Overall, all four metrics show a clear performance enhancement from $\widehat{\boldsymbol{x}}_{9}$ to $\widehat{\boldsymbol{x}}_{0}$, demonstrating the effectiveness of the dynamic optimization strategy.

\begin{table}[!t]
\centering
\caption{Ablation study on the influence of the timestep $T$. The best result of each metric is shown in \textbf{bold}.}
\label{Ablation_timestep}
\renewcommand{\arraystretch}{1.25}
\renewcommand{\tabcolsep}{1.6mm}
{%
\begin{tabular}{c||c|cccc}
\toprule
 Timestep  & Speed  & \multicolumn{4}{c}{EORSSD~\citep{2021DAFNet}}   \\
 \cline{3-6}
  $T$ & (fps) & $S_{\alpha}\uparrow$ & $F_{\beta}^{\rm{max}}\uparrow$ & $E_{\xi}^{\rm{max}}\uparrow$ & $ \mathcal{M}\downarrow$\\
\midrule
1  & 24.3 & .9395  & .8974  & .9821  & .0054  \\
5  & 7.9 & .9455  & .9029  & .9845  & \textbf{.0044}  \\
10 \textbf{(Ours)}  & 4.0 &  \textbf{.9461}  & \textbf{.9033}  & \textbf{.9861}  & \textbf{.0044}    \\ 
15  & 2.8 & .9419  & .8977  & .9818  & .0045  \\
20  & 2.1 & .9413  & .8967  & .9832  & .0049  \\
50  & 0.9 & .9412  & .8989  & .9844  & \textbf{.0044}  \\
100 & 0.4 & .9419  & .8965  & .9839  & .0049  \\

\bottomrule
\end{tabular}%
}
\end{table}

To intuitively verify the effectiveness of the dynamic optimization strategy, we provide a timestep-wise analysis from $\widehat{\boldsymbol{x}}_{9}$ to $\widehat{\boldsymbol{x}}_{0}$ in Fig.~\ref{fig:VisualDynamicOptimization}.
By decomposing the prediction errors into false positives (red) and false negatives (green), we observe a clear progressive correction process.
First, the strategy demonstrates a remarkable capability in background noise suppression, \ie false positive reduction. 
As shown in the first and second cases, the initial saliency maps ($\widehat{\boldsymbol{x}}_{9}$) often contain red regions caused by complex background interference, such as riverside textures or shadows. 
Through the iterative optimization, these false positives are gradually pruned and nearly disappear by $\widehat{\boldsymbol{x}}_{4}$.
Second, the progressive recovery of object details (\ie false negative reduction) is highly evident. 
In the third and fourth cases, IPDiff initially fails to capture the internal structure (marked in green). 
However, as the timestep $t$ evolves, the green areas are systematically replaced by white pixels (\ie true positives). 
This proves that the dynamic optimization strategy allows IPDiff to recover the missing parts and maintain internal consistency within objects.
Finally, the strategy can sharpen object boundaries. 
Across all cases, particularly the slender road in the last case, the coarse and fragmented predictions in the early stages ($\widehat{\boldsymbol{x}}_{9}$-$\widehat{\boldsymbol{x}}_{7}$) are refined into relatively crisp and continuous boundaries in $\widehat{\boldsymbol{x}}_{0}$.
The above cases demonstrate that our dynamic optimization strategy effectively polishes the saliency maps to achieve higher accuracy.

\textit{7) Influence of the Timestep $T$:}
The timestep $T$ is a critical hyperparameter of our IPDiff.
To evaluate the influence of the timestep $T$, we additionally conduct dedicated experiments using $T \in \{1, 5, 15, 20, 50, 100\}$, while keeping all other settings constant.
We report the experimental results and inference speed (fps) in Tab.~\ref{Ablation_timestep}.
We observe that increasing $T$ from 1 to 10 yields consistent improvements across all metrics.
For instance, $S_\alpha$ improves from 0.9395 to 0.9461. 
Although the inference speed decreases from 24.3 fps to 4.0 fps, $T=10$ provides a high-quality prediction while maintaining a decent inference speed.
Notably, the performance does not increase linearly with the number of timesteps. 
When $T$ exceeds 10, the metrics, such as \eg $F_{\beta}^{\text{max}}$ and $E_{\xi}^{\text{max}}$, begin to saturate or even fluctuate slightly, while the computational cost increases substantially (\eg the inference speed drops to merely 0.4 fps at $T=100$).
This indicates that our diffusion-based dynamic optimization strategy does not perform better with a longer timestep, and its error correction capability reaches a plateau rather than improving indefinitely with additional timesteps.
Therefore, based on these analyses, we select $T=10$ as the default setting for IPDiff.

\begin{table*}[t!]
  \centering
  \renewcommand{\arraystretch}{1.25}
  \renewcommand{\tabcolsep}{0.8mm}
  \caption{Generalization ability evaluation with state-of-the-art representative methods on the EORSSD dateset, the 360-SOD dateset, and the 360-SSOD dateset.
  The first dataset is used for cross-dataset testing, while the last two datasets are 360$^\circ$ omnidirectional image SOD datasets used for zero-shot testing.
  The best result of each metric is shown in \textbf{bold}.}
\label{table:GeneralizatioResults}
  
\resizebox{1\textwidth}{!}{
\begin{tabular}{r|cccc|cccc|cccc|cccc}
\midrule[1pt]    

 \multirow{3}{*}{\normalsize{Methods}}
 & \multicolumn{8}{c|}{Cross-dataset Testing} 
 & \multicolumn{8}{c}{Zero-shot Testing}   \\
  \cline{2-8} \cline{9-17}

 & \multicolumn{4}{c|}{ORSI-4199$\rightarrow$EORSSD} 
 & \multicolumn{4}{c|}{ORSSD$\rightarrow$EORSSD} 
 & \multicolumn{4}{c|}{ORSSD$\rightarrow$360-SOD} 
 & \multicolumn{4}{c}{ORSSD$\rightarrow$360-SSOD} \\

 \cline{2-5} \cline{6-9} \cline{10-13} \cline{14-17}
       & $S_{\alpha}\uparrow$ & $F_{\beta}^{\rm{max}}\uparrow$ & $E_{\xi}^{\rm{max}}\uparrow$ & $ \mathcal{M}\downarrow$
   	          & $S_{\alpha}\uparrow$ & $F_{\beta}^{\rm{max}}\uparrow$ & $E_{\xi}^{\rm{max}}\uparrow$ & $ \mathcal{M}\downarrow$
	          & $S_{\alpha}\uparrow$ & $F_{\beta}^{\rm{max}}\uparrow$ & $E_{\xi}^{\rm{max}}\uparrow$ & $ \mathcal{M}\downarrow$
            & $S_{\alpha}\uparrow$ & $F_{\beta}^{\rm{max}}\uparrow$ & $E_{\xi}^{\rm{max}}\uparrow$ & $ \mathcal{M}\downarrow$\\

\midrule[1pt]

ACCoNet$_{23}$ 	 & .8249 & .7375 & .8565  & .0640 
									   & .8871 & .8117 & .9229  & .0128  
									   & .5281 & .2121  & .6402 & .0860 
                                       & .5099 & .1957  & .6039 & .0987  \\
GeleNet$_{23}$ 	 & .8716 & .7883 & .9097  & .0258 
									   & .8927 & .8205 & .9374  & .0154 
									   & .5639 & .3002  & .6808 & .0764 
                                       & .5247 & .2458  & .6061 & .0889  \\

TSCNet$_{24}$   & .8307 & .7902 & .8974  & .0658 
									   & .8823 & .8112 & .9233  & .0215  
									   & .4845 & .1771  & .5971 & .1455 
                                       & .4717 & .1529  & .5849 & .1366  \\

DPU-Former$_{25}$  & .8558 & .8058 & .9068  & .0445 
									   & .8923 & \textbf{.8341} & \textbf{.9415}  & .0134  
									   & .5423 & .2814  & .6313 & .1393 
                                       & .5242 & .2627  & .6248 & .1297  \\

\hline
\hline
                                        
\textbf{IPDiff (Ours)}				  & \textbf{.8928} & \textbf{.8284} & \textbf{.9354} & \textbf{.0146}
									   & \textbf{.8960} & .8338 & .9411 & \textbf{.0113}  
									   & \textbf{.6026} & \textbf{.3775} & \textbf{.7366} & \textbf{.0638}  
                                       & \textbf{.5544} & \textbf{.3177} & \textbf{.6933} & \textbf{.0879}  \\
                            
\bottomrule[1pt]
\end{tabular}
} 
\end{table*}

\textit{8) Generalization Ability Evaluation:}
To further evaluate the generalization and robustness of our IPDiff, we conduct cross-dataset testing and zero-shot testing, as reported in Tab.~\ref{table:GeneralizatioResults}.
We compare our IPDiff with several representative methods, including ACCoNet, GeleNet, TSCNet, and DPU-Former.

For cross-dataset testing, we train IPDiff on one ORSI-SOD dataset and test on another, \ie ORSI-4199$\to$EORSSD and ORSSD$\to$EORSSD. 
Compared with state-of-the-art competitors, our IPDiff achieves superior performance. 
Notably, in the ORSI-4199$\to$EORSSD task, IPDiff outperforms the second-best method, GeleNet, by 2.12\% in $S_\alpha$ and 4.01\% in $F_\beta^{\max}$. 
This demonstrates that IPDiff effectively captures invariant structural features across different optical remote sensing platforms and sensors.
We also observe that IPDiff and DPU-Former each have their own advantages in the ORSSD$\to$EORSSD task. 
This is mainly because the EORSSD dataset is a direct extension of the ORSSD dataset, resulting in a highly similar data distribution.
This indicates that while both models fit well to in-distribution data, the unique advantage of IPDiff lies in its superior generalization to more challenging and unseen senses.

We further challenge IPDiff by performing zero-shot testing on 360$^\circ$ omnidirectional images.
Specifically, we directly applied the weights trained on the ORSSD dataset to the 360-SOD dataset~\citep{360-SOD} and the 360-SSOD dataset~\citep{360-SSOD}, which are two 360$^\circ$ omnidirectional image SOD datasets, \ie ORSSD$\to$360-SOD and ORSSD$\to$360-SSOD.
This is a rigorous task due to the significant domain gap between top-down remote sensing views and panoramic perspectives. 
Remarkably, IPDiff maintains high performance, significantly surpassing all compared methods. 
For instance, on the 360-SOD dataset, IPDiff achieves a 9.61\% improvement in $F_\beta^{\max}$ over DPU-Former. 
These results indicate that our information perturbation and multi-prior guidance enable the IPDiff to learn the intrinsic manifold of saliency beyond specific imaging domains.

\section{Conclusion}
\label{sec:con}
In this paper, we propose IPDiff, a novel diffusion-driven ORSI-SOD framework.
IPDiff follows a unique dynamic optimization strategy to iteratively optimize saliency maps with the dynamically changing timestep in the testing phase.
In IPDiff, we formulate ORSI-SOD as a conditional diffusion problem.
Specifically, we introduce a prior network to extract the saliency prior and the hierarchical priors from ORSIs as conditional priors.
In the prior network, we employ IRAMs to adaptively reconstruct information in the spectral domain to increase the useful content of multiple priors.
Then, we inject the above priors into the carefully designed denoising network.
In the denoising network, the noisy mask is combined with the saliency prior and the hierarchical priors successively to absorb specific position, detail, and semantic information of salient objects.
IPMs are equipped in the denoising network in the training phase to stabilize feature representation and provide anti-interference capability.
As the timestep dynamically changes, the noisy mask is iteratively optimized to recover a clear saliency map.
Comprehensive experiments, including quantitative and qualitative comparisons and ablation studies, demonstrate the leading position of IPDiff in the ORSI-SOD community and the effectiveness of its components.

\section*{Data Availability Statement}
The datasets used in this work are all publicly available. ORSSD is available at \url{https://li-chongyi.github.io/proj\_optical\_saliency.html}. EORSSD is available at \url{https://github.com/rmcong/DAFNet\_TIP20}. ORSI-4199 is available at \url{https://github.com/wchao1213/ORSI-SOD}.

\bibliographystyle{spbasic}

\bibliography{IJCVRef}

\end{document}